\documentclass{article}

\usepackage{PRIMEarxiv}

\usepackage[utf8]{inputenc} 
\usepackage[T1]{fontenc}    
\usepackage{hyperref}       
\usepackage{url}            
\usepackage{booktabs}       
\usepackage{amsfonts}       
\usepackage{nicefrac}       
\usepackage{microtype}      
\usepackage{lipsum}
\usepackage{fancyhdr}       
\usepackage{graphicx}       
\graphicspath{{media/}}     

\usepackage{caption}
\usepackage{float}
\usepackage{glossaries}
\usepackage{subcaption}

\DeclareMathOperator{\rmse}{RMSE}
\DeclareMathOperator{\brmse}{BRMSE}
\DeclareMathOperator{\drs}{DRS}
\DeclareMathOperator{\prs}{PRS}
\DeclareMathOperator{\tars}{TARS}
\DeclareMathOperator{\clip}{Clip}

\newacronym{cnn}{CNN}{Convolutional Neural Network}
\newacronym{dl}{DL}{Deep Learning}
\newacronym{drs}{DRS}{Deformation Robustness Score}
\newacronym{dwd}{DWD}{German Meteorological Service}
\newacronym{entso-e}{ENTSO-E}{European Network of Transmission System Operators}
\newacronym{lstm}{LSTM}{Long Short-Term Memory}
\newacronym{mae}{MAE}{Mean Absolute Error}
\newacronym{ml}{ML}{Machine Learning}
\newacronym{mse}{MSE}{Mean Square Error}
\newacronym{pgd}{PGD}{Projected Gradient Descent}
\newacronym{prs}{PRS}{Performance Robustness Score}
\newacronym{rmse}{RMSE}{Root Mean Square Error}
\newacronym{brmse}{BRMSE}{Bounded Root Mean Square Error}
\newacronym{tars}{TARS}{Total Adversarial Robustness Score}
  
\title{Targeted Adversarial Attacks on Wind Power Forecasts
}

\author{
  René Heinrich\textsuperscript{1,2,*}, Christoph Scholz\textsuperscript{1,2},
  Stephan Vogt\textsuperscript{2},
  Malte Lehna\textsuperscript{1,2} \\ \\
  \textsuperscript{1} Fraunhofer Institute for Energy Economics and Energy System Technology (IEE) \\ 
  Joseph-Beuys-Straße 8, 34117 Kassel, Germany \\ \\
  \textsuperscript{2} Intelligent Embedded Systems (IES), University of Kassel \\
  M\"onchebergstraße 19, 34127 Kassel, Germany \\ \\
  * rene.heinrich@iee.fraunhofer.de \\
}

\begin{document}
\maketitle

\begin{abstract}
In recent years, researchers proposed a variety of deep learning models for wind power forecasting.
These models predict the wind power generation of wind farms or entire regions more accurately than traditional machine learning algorithms or physical models.
However, latest research has shown that deep learning models can often be manipulated by adversarial attacks.
Since wind power forecasts are essential for the stability of modern power systems, it is important to protect them from this threat. \\
In this work, we investigate the vulnerability of two different forecasting models to targeted, semi-targeted, and untargeted adversarial attacks. 
We consider a \gls{lstm} network for predicting the power generation of individual wind farms and a \gls{cnn} for forecasting the wind power generation throughout Germany.
Moreover, we propose the \gls{tars}, an evaluation metric for quantifying the robustness of regression models to targeted and semi-targeted adversarial attacks. 
It assesses the impact of attacks on the model's performance, as well as the extent to which the attacker's goal was achieved, by assigning a score between 0 (very vulnerable) and 1 (very robust).
In our experiments, the \gls{lstm} forecasting model was fairly robust and achieved a \gls{tars} value of over 0.78 for all adversarial attacks investigated. 
The \gls{cnn} forecasting model only achieved \gls{tars} values below 0.10 when trained ordinarily, and was thus very vulnerable.
Yet, its robustness could be significantly improved by adversarial training, which always resulted in a \gls{tars} above 0.46.
\end{abstract}


\section{Introduction}
\label{sec:introduction}
\begin{figure}[H]
    \centering
    \includegraphics[trim={0cm 0cm 0cm 0cm}, clip, width=\textwidth]{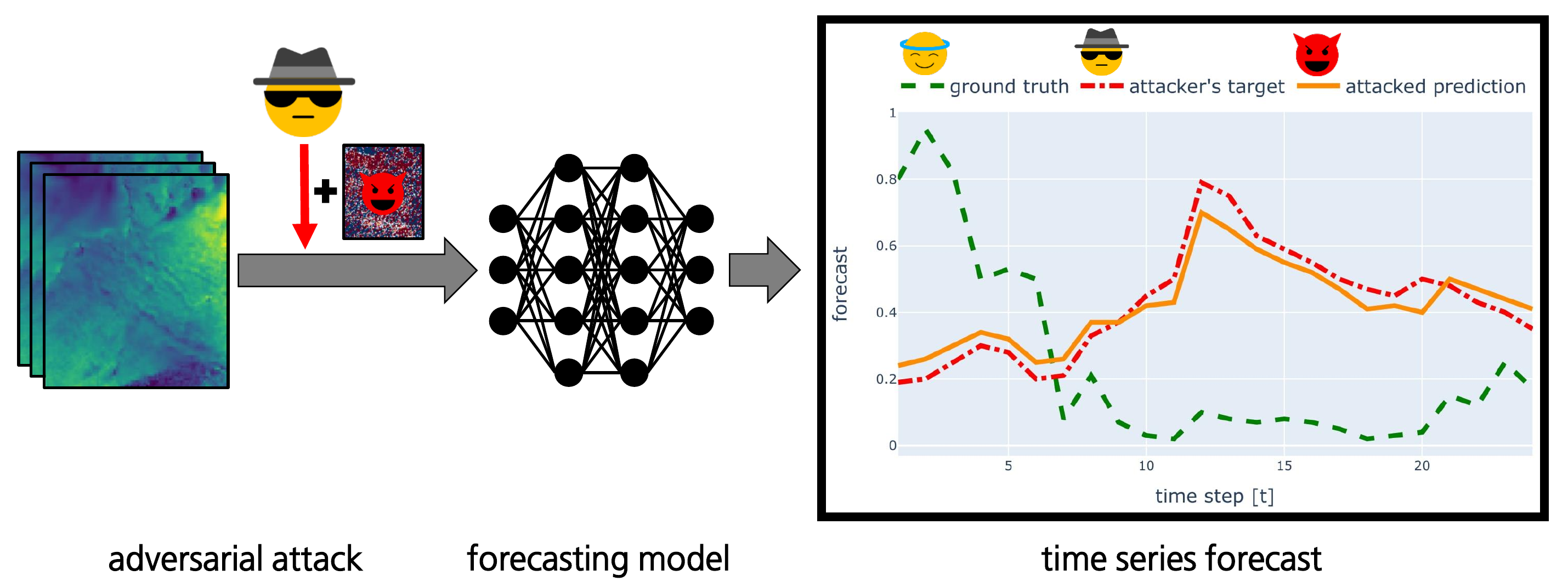}
    \caption{Illustration of a targeted adversarial attack on a time series forecasting model.
    The adversary manipulates the input data by adding a small perturbation. This perturbation causes the model's prediction (solid) to no longer approximate the ground truth (dashed), but to follow a particular forecast pattern (dash-dotted) defined by the attacker
    }
    \label{fig:illustration_targeted_attack}
\end{figure}

Renewable energy forecasting has a significant impact on the planning, management, and operation of power systems \cite{wang2019review}.
Grid operators and power plants require accurate forecasts of renewable energy output to ensure grid reliability and permanency, and to reduce the risks and costs of energy markets and power systems \cite{alkhayat2021review}.
Over the past few years, the share of renewable energies in the electricity mix has risen steadily.
For example, the total installed wind energy capacity in Germany increased from 26.9 gigawatts in 2010 to 63.9 gigawatts in 2021 \cite{deutschland2022renewable}.
Moreover, wind energy already covered about 20 percent of the German gross electricity consumption in 2021, making it the most important energy carrier in the German electricity mix.
This development poses a challenge for energy providers. 
Wind power generation is difficult to predict due to the randomness, volatility, and intermittency of wind.
Improving the accuracy of wind power forecasts is therefore of high importance. \\
In recent years, \gls{dl} methods have proven to be particularly feasible and effective for accurate renewable energy forecasting \cite{wang2019review,alkhayat2021review,aslam2021survey}.
Nevertheless, power systems are a critical infrastructure that can be targeted by criminal, terrorist, or military attacks.
Hence, not only the accuracy of wind power forecasts is relevant, but also their attack resistance.
Latest research has shown that \gls{dl} methods are often vulnerable to adversarial attacks \cite{szegedy2013intriguing,goodfellow2014explaining}.
The use of \gls{dl} thus poses dangers and opens up new attack opportunities for assailants.
Adversarial attacks slightly perturb the input data of \gls{ml} models to falsify their predictions.
In particular, \gls{dl} algorithms that obtain input data from safety-critical interfaces are exposed to this threat.
Wind power forecasting models often use satellite imagery or weather forecasts as input features. 
Such data frequently comes from publicly available data sources which can be corrupted by hackers. 
Even data sources that are not public can become the target of attacks. 
For example, there is a risk that energy data markets \cite{goncalves2020towards} will be abused by attackers in the future. Attackers could use these markets to inject tampered data into an \gls{ml} application and thereby manipulate its predictions. 
If such manipulations remain undetected and if forecasting models are not adequately protected, the consequences could be fatal. 
Attacks on wind power forecasts could compromise forecast quality, resulting in high costs for energy consumers and energy providers.
Even worse, attackers could also manipulate the forecasts to gain economic advantages or destabilize energy systems. \\ \\
Consequently, there is a growing interest among researchers to study the effects of adversarial attacks in the context of time series data.
In particular, the vulnerability of \gls{dl} methods for time series classification has been studied by various researchers \cite{fawaz2019adversarial,abdu2020detecting,rathore2020untargeted}. They considered adversarial attacks such as the Fast Gradient Sign Method \cite{goodfellow2014explaining} and the Basic Iterative Method \cite{kurakin2018adversarial} to cause misclassification of time series data.
More advanced techniques such as the Adversarial Transformation Network \cite{karim2020adversarial,harford2020adversarial} have also been proposed for this purpose.
However, adversarial attacks on \gls{ml} algorithms are also highly relevant for regression tasks such as time series forecasting \cite{alfeld2016data}.
With respect to \gls{dl} approaches, \cite{nguyen2018adversarial} examined the impact of adversarial attacks on regression neural networks and proposed a stability-inducing, regularization-based defense against these attacks. 
Nevertheless, adversarial attacks for regression tasks still require additional research, as the number of contributions on this topic is yet relatively limited. \\
With the rising adoption of \gls{dl} in the power industry, the analysis and detection of adversarial attacks is becoming a growing concern. 
Since energy systems are critical infrastructures, the security of \gls{dl} algorithms in this domain is of particular importance. 
According to \cite{richter2022artificial}, the \gls{dl} models deployed in this field can become targets of attacks across the entire value chain.
In this regard, an important topic of interest is the protection of grid infrastructures and smart grids against adversarial attacks. 
The survey of \cite{cui2020detecting} shows that various papers related to false data injection attacks have already been published in this sector. 
There also exists research that investigates the threat of adversarial attacks designed to fool anomaly detection methods \cite{ahmadian2018cyber,sayghe2020evasion}. 
Other papers cover grid-related topics such as utilizing adversarial attacks for the purpose of energy theft in energy management systems \cite{marulli2019adversarial} or attacks on event cause analysis \cite{niazazari2020attack}. 
Another important research direction in the energy domain are adversarial attacks on power forecasts. 
Here, \cite{zhou2019evaluating} have shown that the prediction accuracy of load flow forecasts can be degraded by stealthy adversarial attacks. 
Further, \cite{chen2019exploiting} have analyzed how load flow forecasts can be biased in a direction advantageous to the attacker. 
Still other researchers have focused on attacks against renewables.
For instance, \cite{tang2021adversarial} studied the impact of untargeted adversarial attacks on solar power forecasts. \\
In this work, the focus is on wind power forecasting, due to its rising importance in power systems. 
Recently, \gls{dl} models have been increasingly proposed by researchers for this task \cite{alkhayat2021review,wu2022comprehensive}.
However, very little research has been done on the robustness of these models to adversarial attacks. 
A notable contribution was made by \cite{zhang2020robustness}, who approached the problem of false data injection attacks from a technical point of view. 
In doing so, they examined the impact of untargeted adversarial attacks on a variety of regression models, including support vector machines, fully connected neural networks, and quantile regression neural networks. 
In contrast to previous studies, the focus of this work is to investigate targeted adversarial attacks on \gls{dl} models for wind power forecasting. 
The goal of targeted adversarial attacks is to manipulate the forecasting model in such a way that the predicted values follow a specific forecast pattern desired by the attacker, see Fig. \ref{fig:illustration_targeted_attack}. \\
As discussed previously, only untargeted and semi-targeted attacks on \gls{dl}-based forecasting models have been studied so far. 
In the case of wind power forecasts, however, targeted adversarial attacks pose a much greater threat. 
Such attacks give assailants the opportunity to specifically influence forecast behavior. Thus, they are able to affect energy markets or disrupt grid operations. 
Especially in regression tasks, evaluating the success of targeted adversarial attacks is non-trivial.
Therefore, it is important to have appropriate evaluation metrics for assessing the robustness of models to such attacks.
In this work, we address these problems and offer the following contributions:
\begin{enumerate}
    \item[(C1)] We propose a taxonomy for adversarial attacks in the regression setting that categorizes them into untargeted, semi-targeted, and targeted attacks.
    \item[(C2)] We present an evaluation metric for assessing the robustness of regression models to targeted and semi-targeted adversarial attacks. This evaluation metric measures not only the impact of the attacks on the performance of the model, but also the extent to which the attacker's goal was achieved.
    \item[(C3)] 
    We investigate the robustness of two different \gls{dl} models for wind power forecasting, each with its own use case. 
    We find that \gls{cnn} models for predicting the wind power generation throughout Germany based on wind speed forecasts in the form of weather maps are very susceptible to adversarial attacks, whereas \gls{lstm} models for predicting the power generation of wind farms based on wind speed forecasts in the form of time series are fairly robust.
    \item[(C4)] 
    We examine the effects of adversarial training and show that it significantly increases the robustness of the \gls{cnn} forecasting model, while having only a small effect on the robustness of the \gls{lstm} forecasting model in the respective applications.
\end{enumerate}
This paper is organized as follows. 
In Section \ref{sec:method}, we present the underlying methodology behind adversarial attacks and adversarial training. 
Moreover, an evaluation metric for quantifying the adversarial robustness of regression models is proposed.
Next, two different \gls{dl}-based wind power forecasting models are investigated in terms of their robustness to adversarial attacks. 
First, the experimental setup is presented in Section \ref{sec:experiments}. 
Subsequently, the results of the study are presented in Section \ref{sec:results}.
In Section  \ref{sec:discussion}, a discussion of the results follows and several directions for future work are pointed out. Finally, we conclude with a summary of this contribution in Section \ref{sec:conclusion}.

\section{Methodology}
\label{sec:method}

\subsection{Adversarial attacks}
\label{subsec:adv_attacks}
Adversarial attacks refer to attacks on \gls{ml} algorithms that perturb the input data in order to manipulate the model's prediction. 
In the process, the attacker modifies the input data slightly and carefully, so that the perturbations remain undetected by humans and anomaly detection methods. 
The techniques for generating adversarial attacks can be taxonomically categorized according to the attacker's goal and the prior knowledge of the attacker \cite{xu2020adversarial}.
Whereas white-box adversarial attacks require complete knowledge about the model architecture and the trained model parameters, gray-box methods assume only limited knowledge of the attacker, e.g., about confidence levels of the model. 
Black-box methods, on the other hand, suppose that the attacker has no knowledge about the underlying model. 
However, it is commonly assumed that the attacker is able to communicate with the model. \\
Regarding the attacker's goal, a distinction is made between untargeted and targeted attacks in classification tasks. 
The goal of targeted attacks is to fool the model into classifying the input as a particular class desired by the adversary. 
In contrast, untargeted attacks simply aim for a misclassification of the perturbed data. 
The exact class predicted by the model is not important. 
For regression tasks, though, the output of \gls{ml} algorithms is not categorical, but represents continuous variables. 
Thus, this categorization of adversarial attacks cannot be simply transferred to regression problems. 

\subsubsection{Goals of adversarial attacks in regression tasks}
\label{subsubsec:goals_regression}
\begin{figure}[h]
  \begin{minipage}[t]{0.32\textwidth}
  \caption{Example of an untargeted adversarial attack. While the original prediction (dotted) approximates the ground truth (dashed) very well, the attacked prediction (solid) deviates strongly from the ground truth}
  \label{fig:taxonomy_untargeted_attack}
  \end{minipage} \hfill
  \begin{minipage}[t]{0.65\textwidth}
   \vspace{0pt}
   \includegraphics[trim={0 0.5cm 0 2cm},clip, width=\textwidth]{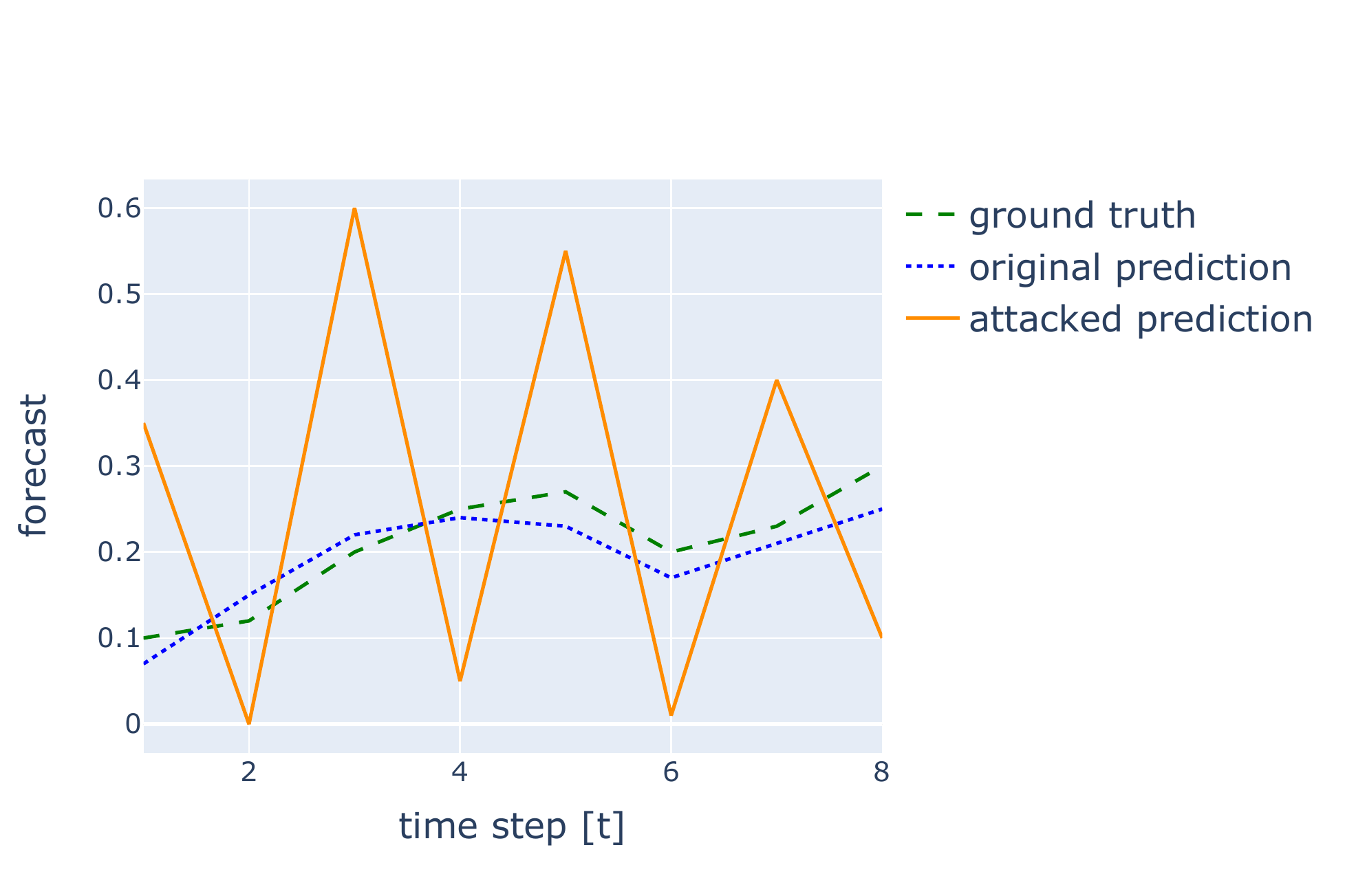}
  \end{minipage}
\end{figure}

\begin{figure}[h]
  \begin{minipage}[t]{0.33\textwidth}
  \caption{Example of a semi-targeted adversarial attack. While the original prediction (dotted) approximates the ground truth (dashed) very well, the attacked prediction (solid) lies in the area defined by the attacker's constraints (dash-dotted)}
  \label{fig:taxonomy_semitargeted_attack}
  \end{minipage} \hfill
  \begin{minipage}[t]{0.64\textwidth}
   \vspace{0pt}
   \includegraphics[trim={0 0.5cm 0 2cm},clip, width=\textwidth]{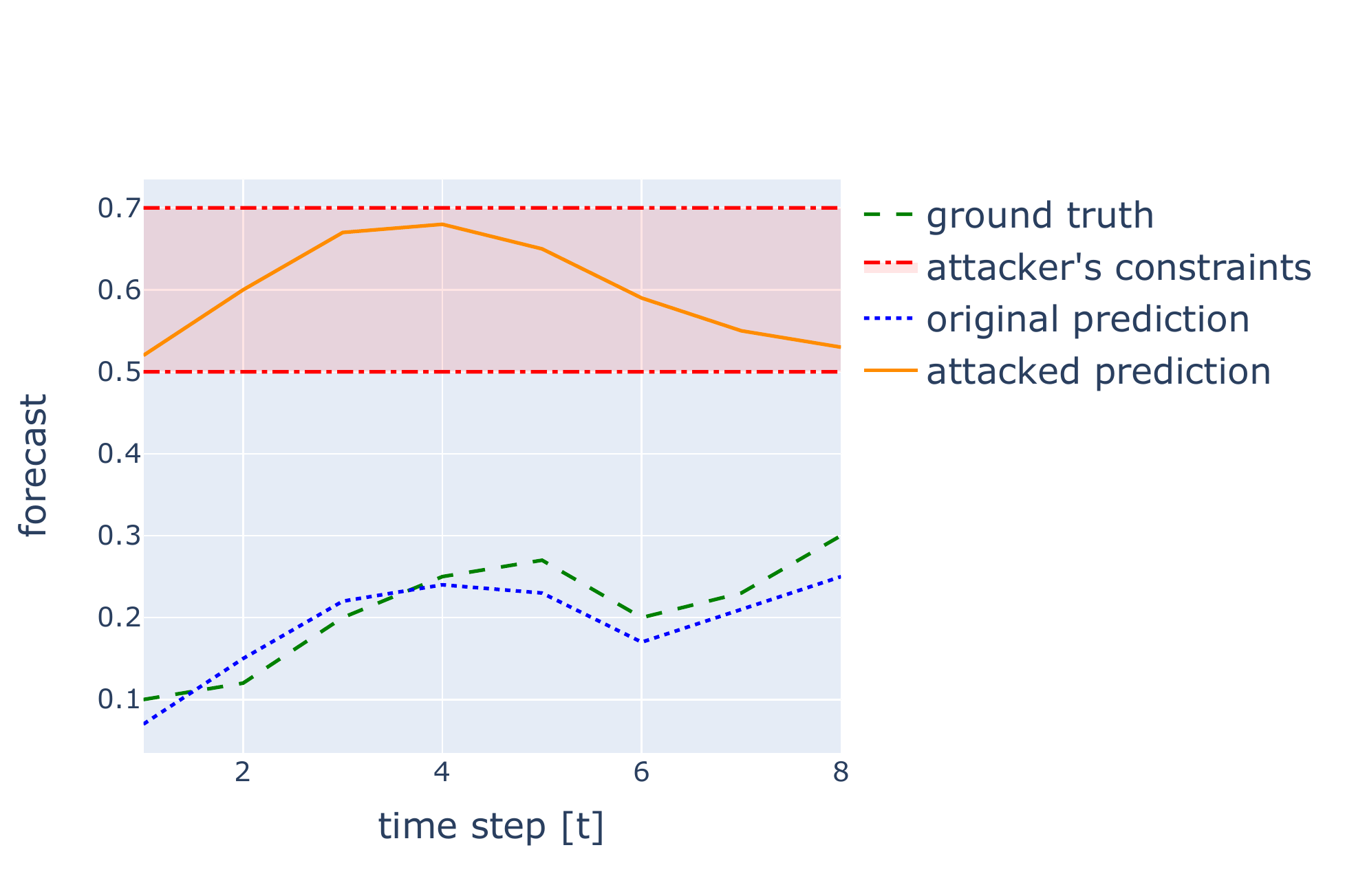}
  \end{minipage}
\end{figure}

\begin{figure}[h]
  \begin{minipage}[t]{0.33\textwidth}
  \caption{Example of a targeted adversarial attack. While the original prediction (dotted) almost matches the ground truth (dashed), the attacked prediction (solid) approximates the attacker's target (dash-dotted)}
  \label{fig:taxonomy_targeted_attack}
  \end{minipage} \hfill
  \begin{minipage}[t]{0.64\textwidth}
   \vspace{0pt}
   \includegraphics[trim={0 0.5cm 0 2cm},clip, width=\textwidth]{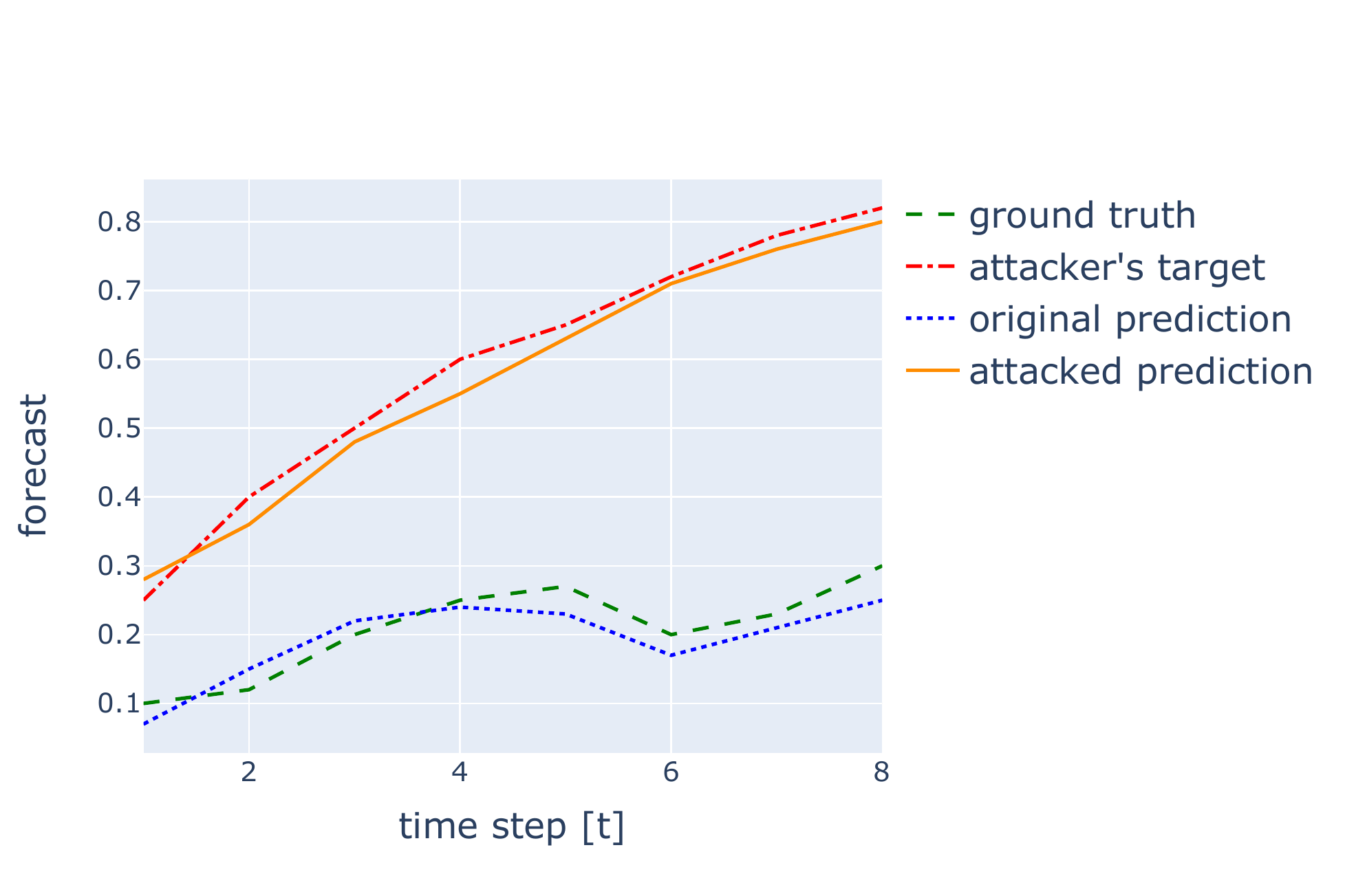}
  \end{minipage}
\end{figure}

As contribution (C1), we propose to taxonomically divide the attacker's goal into three categories in the regression setting: untargeted attacks, semi-targeted attacks, and targeted attacks.
Untargeted attacks attempt to perturb an input data point $x \in \mathbb R^d$ in such a way that the prediction quality of a model $f_{\theta}$, with parameters $\theta \in \mathbb R^p$, is degraded to the maximum in terms of a loss function $\mathcal{L}$. 
The objective that the attacker wants to optimize is as follows:
\begin{equation}
    \max_{\delta  \in \mathcal{S}} \mathcal{L} \left( f_{\theta} (x + \delta), y \right)
\end{equation}
Here, $y \in \mathbb R^n$ is the ground truth value associated with the input data point $x$. The perturbation added to $x$ is denoted by $\delta$, and $\mathcal{S} \subseteq \mathbb R^d$ represents the set of allowed perturbations.
An example of an untargeted adversarial attack on a univariate time series forecast is shown in Figure \ref{fig:taxonomy_untargeted_attack}.
\newline
In the case of untargeted attacks, the attacker has no control over the magnitude of the degradation.
Thus, he risks that the attack will result in an unrealistic prediction that can easily be detected as erroneous. \\
To avoid this, attackers also have the option of launching semi-targeted attacks on regression models. 
We define semi-targeted attacks as perturbations that cause the model's predictions to fall within certain boundaries.
These boundaries are specified by the attacker. 
Thus, the perturbations aim at degrading the model's performance, while satisfying certain constraints:
\begin{equation}
\begin{split}
        \max_{\delta  \in \mathcal{S}} \quad & \mathcal{L} \left( f_{\theta} (x + \delta), y \right) \\
        \text{s.t.} \quad & C_i \left(f_{\theta} (x + \delta) \right) \leq 0 \quad \text{for } i = 1, \dots, k \\
        & C_j \left(f_{\theta} (x + \delta) \right) = 0 \quad \text{for } j = 1, \dots, l
\end{split}
\end{equation}
Here, the inequality constraints $C_i$ and the equality constraints $C_j$ describe the attacker's desired restrictions on the behavior of the manipulated prediction $f_{\theta} (x + \delta)$.
For example, the attacker may attempt to degrade the prediction quality only to a certain degree so that the degradation remains inconspicuous.
Another example are perturbations that cause the prediction to be distorted as much as possible in a certain direction, e.g., to either increase or decrease the predicted values, as was studied by \cite{chen2019exploiting}. 
In this work, we study semi-targeted adversarial attacks with lower and upper bound constraints. 
Here, the attacker specifies a lower bound $a \in \mathbb R^n$ and an upper bound $b \in \mathbb R^n$. 
The attacker then attempts to perturb the input data such that the attacked prediction $\hat y _{adv} = f_{\theta} (x + \delta)$ falls within the region enclosed by the lower and upper bound, i.e., $a_i \leq \hat y _{adv, i} \leq b_i$ holds for all $i = 1, \dots , n$.
In the example in Figure \ref{fig:taxonomy_semitargeted_attack}, the constraints require the prediction $\hat y _{adv}$ to only take values between 0.5 and 0.7.
\newline
Finally, regression models can also be manipulated by attackers in a targeted fashion. 
Targeted attacks try to perturb the input data in such a way that the model's prediction comes as close as possible to an adversarial target $y_{adv} \in \mathbb R^n$.
Thus, the attacker aims for the following optimization objective:
\begin{equation}
    \min_{\delta  \in \mathcal{S}} \mathcal{L} \left( f_{\theta} (x + \delta), y_{adv} \right)
\end{equation}
Depending on the application, different target values may be relevant for the attacker. 
For instance, an attacker could try to manipulate wind power forecasts in order to influence energy markets and gain economic advantages. 
An example of a targeted adversarial attack is shown in Figure \ref{fig:taxonomy_targeted_attack}.
\noindent
In this paper, two methods for generating adversarial attacks are considered. 
The focus is on untargeted, semi-targeted, and targeted adversarial attacks using the \gls{pgd} attack. 
In addition, we also examine untargeted adversarial noise attacks, which are rather weak attacks but serve as a baseline. 
The two methods are described below.

\subsubsection{Adversarial noise attack}
\label{subsubsec:noise_attack}
A very simple form of untargeted adversarial attacks are adversarial noise attacks, which were originally introduced by \cite{rauber2017foolbox}.
Noise attacks are applicable to both classification tasks and regression tasks.
They perturb the input data by adding random noise, commonly Gaussian noise or uniform noise. In the process, the perturbation is normalized and rescaled to the desired size, e.g., with respect to the $L_{\infty}$ norm. In addition, the perturbed samples need to be clipped afterwards so that all values are within the valid lower and upper bounds of the input data \cite{rauber2020fast}.
Noise attacks require no prior knowledge of the model and thus represent black-box attacks. 
In order to increase the success rate of the attack, repeated noise attacks can be used. 
Here, noise is repeatedly sampled, thus generating several candidate noise terms for the attack. 
Then the effects of the different noise terms on the model's performance are evaluated. 
Finally, the noise term that most degrades the model's performance is selected as the perturbation.

\subsubsection{Projected Gradient Descent (PGD) attack}
\label{subsubsec:pgd_attack}
According to \cite{carlini2019evaluating}, by far the most powerful attack algorithms are those that use gradient-based optimization.
They extract a significant amount of information from the model by using the gradients of a loss function to generate adversarial attacks. 
One such optimization-based attack commonly used in the literature is \gls{pgd}, which was originally proposed by \cite{madry2017towards}. 
\gls{pgd} attempts to iteratively improve the perturbation of an input, while always ensuring that the magnitude of the perturbation is within a given boundary. 
To do this, \gls{pgd} exploits the model gradients between the input and an adversarial loss function.
Thus, it is a white-box attack and applicable for untargeted, semi-targeted as well as targeted attacks. \\
In the case of untargeted attacks, \gls{pgd} attempts to maximize the deviation between the model's prediction and the ground truth \cite{kurakin2018adversarial}:
\begin{equation}
    x_{adv}^{\left(0\right)} = x, \quad x_{adv}^{\left(t+1\right)} = \clip_{x, \epsilon} \left\{ x_{adv}^{\left(t\right)} + \alpha \text{ sign} \left( \nabla _{x_{adv}^{\left(t\right)}} \mathcal{L} \left( f_{\theta} \left( x_{adv}^{\left(t\right)} \right), y \right) \right) \right\}
\end{equation}
On the other hand, in targeted attacks, \gls{pgd} tries to minimize the mismatch between the model's prediction and the attacker's target \cite{kurakin2018adversarial}:
\begin{equation}
    x_{adv}^{\left(0\right)} = x, \quad x_{adv}^{\left(t+1\right)} = \clip_{x, \epsilon} \left\{ x_{adv}^{\left(t\right)} - \alpha \text{ sign} \left( \nabla _{x_{adv}^{\left(t\right)}} \mathcal{L} \left( f_{\theta} \left( x_{adv}^{\left(t\right)} \right), y_{adv} \right) \right) \right\}
\end{equation}
Here $\alpha$ is the update size per step and $x_{adv}^{\left(t\right)}$ denotes the perturbed input after the $t^{th}$ optimization step.
Feature-wise clipping of the perturbed input using the $\clip_{x, \epsilon}$ function ensures that the result is in the $\epsilon$-neighborhood of the original input $x$, with respect to the $L_{\infty}$ norm. The parameter $\epsilon$ corresponds to the maximum perturbation magnitude specified by the attacker.
It should be noted that \cite{madry2017towards} proposed to add a random initialization to this algorithm. 
However, in the following experiments we always use \gls{pgd} without a random initialization, since it did not have a significant effect on the results in preliminary tests. \\
For applying \gls{pgd} to semi-targeted attacks, we propose to add a weighted penalty term to the loss function, which penalizes the violation of the attacker's constraints. In the case of semi-targeted attacks with lower and upper bound constraints, \gls{pgd} then attempts to maximize the mismatch between the model's prediction and the ground truth, while at the same time minimizing the deviation between the prediction and the area enclosed by the lower and upper bounds:
\begin{equation}
    \begin{split}
    & x_{adv}^{\left(0\right)} = x, \\
    & x_{adv}^{\left(t+1\right)} = \clip_{x, \epsilon} \left\{ x_{adv}^{\left(t\right)} + \alpha \text{ sign} \left( \nabla _{x_{adv}^{\left(t\right)}} \mathcal{L}_{\lambda} \left( f_{\theta} \left( x_{adv}^{\left(t\right)} \right), y, a, b \right) \right) \right\}, \\
    & \mathcal{L}_{\lambda} \left( f_{\theta} \left( x_{adv}^{\left(t\right)} \right), y, a, b \right) = \mathcal{L} \left( f_{\theta} \left( x_{adv}^{\left(t\right)} \right), y \right) - \lambda \cdot \mathcal{L} _{[a,b]} \left( f_{\theta} \left( x_{adv}^{\left(t\right)} \right) \right)
    \end{split}
\end{equation}
Here, $\mathcal{L} _{[a,b]} \left( f_{\theta} \left( x_{adv}^{\left(t\right)} \right) \right)$ is a loss function that serves as the penalty term. It measures the degree of deviation between the prediction and the area enclosed by the lower bound $a$ and the upper bound $b$. The parameter $\lambda$ is the corresponding penalty weight, which was always chosen as 1000 in this work.

\subsection{Adversarial training}
\label{subsec:adv_defenses}
Several techniques exist to protect \gls{ml} algorithms from adversarial attacks \cite{qiu2019review,xu2020adversarial,akhtar2021advances}. 
For example, perturbed data points can be identified and eliminated at an early stage using detection methods \cite{metzen2017detecting}. 
Another approach is to increase a model's robustness.
A robust model is characterized by the fact that it is stable to small perturbations of its inputs \cite{szegedy2013intriguing}.
In a regression setting, this means that minor changes in the input do not lead to significant changes in the model's prediction.
A commonly used technique in the literature is to increase the robustness of a model by adversarial training \cite{goodfellow2014explaining}. 
During adversarial training, the model is trained on perturbed training data. Thus, it automatically becomes more robust to the type of adversarial attacks that were used to generate the perturbations in the training phase.
In each training iteration, the perturbed data points are newly generated from the original training data. 
This ensures that the perturbations are specifically tailored to the model weights of each training iteration.  
Then the model weights $\theta \in \mathbb{R}^p$ are selected by solving the following optimization problem \cite{madry2017towards}:
\begin{equation}
    \min _{\theta} \\\ \mathbb E _{\left( x, y \right) \sim \mathcal D} \left[  \max _{\delta \in \mathcal{S}} \mathcal{L} \left( f_{\theta} \left( x + \delta \right), y \right) \right]
\end{equation}
Here, $\left( x, y \right) \sim \mathcal D$ represents training data sampled from the underlying data distribution $\mathcal D$.
The inner maximization problem is to find the worst-case perturbations for the given model weights, which can be approximately solved by generating adversarial attacks with the \gls{pgd} attack \cite{madry2017towards}. On the other hand, the outer minimization consists in training a model that is robust to these worst-case perturbations. This can be solved by the standard training procedure.

\subsection{Adversarial robustness scores}
\label{subsec:robustness_eval}
In order to evaluate the security of \gls{dl} models, it is essential to quantify their robustness to adversarial attacks. In classification tasks, the success of an attack can be measured quite easily using the model accuracy or the attack success rate \cite{carlini2019evaluating}. 
However, assessing the robustness of regression models is non-trivial, especially in the case of targeted and semi-targeted attacks.
Therefore, as contribution (C2), we present below an evaluation metric for quantifying the robustness of regression models to targeted adversarial attacks and semi-targeted adversarial attacks with lower and upper bound constraints.
From the attacker's perspective, the success of a targeted attack can be measured by the deviation between the model's prediction and the adversarial target.
In the case of semi-targeted attacks, it is important for the attacker that the prediction satisfies his constraints.
But from the victim's point of view, this does not cover all possible harms.
An attack may be unsuccessful for the attacker because the model's prediction is still far from the adversarial target or does not satisfy the attacker's constraints. 
But if the attack significantly degrades the model's performance, it still has a considerable lack of robustness.
Therefore, we propose an evaluation metric to quantify the robustness of regression models specifically for targeted and semi-targeted attacks. \\
In the following, we use the \gls{rmse} to measure the deviation between a model's prediction $\hat y = f_{\theta} \left( x \right) \in \mathbb{R}^n$  and the associated ground truth $y \in \mathbb{R}^n$:
\begin{equation}
    \rmse \left(\hat y, y \right) = \left( \frac{1}{n} \sum_{i=1}^{n} \left( \hat y_i - y_i \right) ^2 \right) ^{\frac{1}{2}}
    \label{eq:rmse}
\end{equation}
The \gls{rmse} has the benefit of penalizing large errors more.
However, it is possible to replace the \gls{rmse} in the scores defined below (DRS, PRS, and TARS) with any other non-negative cost function $\mathcal L$. For example, the \gls{mse} or \gls{mae} are also very common cost functions for regression problems. \\
To quantify the extent to which a prediction $\hat y \in \mathbb{R}^n$ satisfies the lower and upper bound constraints of a semi-targeted attack, we define the following variation of the \gls{rmse}, the \gls{brmse}:
\begin{equation}
    \brmse _{[a,b]} \left(\hat y \right) = \left( \frac{1}{n} \sum_{i=1}^{n} \left( \chi _{\{\hat y_i < a_i\}} \cdot  \left( \hat y_i - a_i \right) ^2 + \chi _{\{ b_i < \hat y_i \}} \cdot \left( \hat y_i - b_i \right) ^2 \right) \right) ^{\frac{1}{2}}
\end{equation}
Here $a \in \mathbb{R}^n$ denotes the lower bound, $b \in \mathbb{R}^n$ the upper bound and $\chi$ the indicator function\footnote{The indicator function $\chi _{\{ x < y \}}$ takes the value 1 if $x < y$ holds and the value 0 if $x \geq y$.}. If a prediction $\hat y$ satisfies the constraints, i.e., if $a_i \leq \hat y_i \leq b_i$ holds for all $i = 1, \dots, n$, then the $\brmse _{[a,b]}$ is zero.
If an element $\hat y_i$ of the prediction is below the lower bound, i.e. if $\hat y_i < a_i$ holds, the $\brmse _{[a,b]}$ accounts only for the deviation between $\hat y_i$ and $a_i$. 
On the other hand, if an element $\hat y_i$ is above the upper bound, i.e. if $\hat y_i > b_i$ holds, the $\brmse _{[a,b]}$ only considers the deviation between $\hat y_i$ and $b_i$. \\
The proposed score for evaluating the robustness to targeted and semi-targeted attacks is composed of two subscores.
These subscores respectively measure the robustness of the model's performance and its robustness to prediction deformations.
The scores are described in more detail below.

\subsubsection{Performance robustness}
\label{subsubsec:performance_robustness}
The first score is the \gls{prs}. 
The \gls{prs} measures how severely a model's performance deteriorates relative to its original performance when under attack:
\begin{equation}
    \prs \left( \hat{y}, \hat{y}_{adv}, y \right) = \min \left( \exp \left( 1 - \frac{\rmse \left(\hat{y}_{adv}, y \right)}{\rmse \left(\hat{y}, y \right) + \gamma}\right), 1 \right)
\end{equation}
Here, $\gamma$ is a small constant value to avoid dividing by zero. In the following we always select $\gamma = 1 \cdot 10^{-10}$.
The \gls{prs} ranges from 0 to 1. 
If the deviation between the model's prediction and the ground truth remains unchanged during the attack or even decreases, the attack has no negative impact on the model's performance. 
In this case, the performance is considered robust to the attack and the \gls{prs} takes the value 1. 
However, if $\rmse \left(\hat y_{adv}, y \right)$ increases relative to $\rmse \left(\hat y, y \right)$, the \gls{prs} converges to zero and the performance robustness decreases exponentially, see Fig. \ref{fig:prs} in Appendix \ref{secA3:robustness_scores}. 

\subsubsection{Deformation robustness}
\label{subsubsec:deformation_robustness}
We define the \gls{drs} to quantify the success of an attacker in case of targeted and semi-targeted attacks. 
For targeted attacks, the \gls{drs} measures how close a model's prediction moves towards the adversarial target due to an attack:
\begin{equation}
    \drs \left( \hat{y}, \hat{y}_{adv}, y_{adv} \right) = \min \left( \exp \left( 1 - \frac{\rmse \left(\hat{y}, y_{adv}\right)}{\rmse \left(\hat{y}_{adv}, y_{adv}\right) + \gamma}\right), 1 \right)
\end{equation}
The \gls{drs} also ranges from 0 to 1.
If the \gls{drs} is equal to 1, the attack has failed from the attacker's point of view. 
This is the case if the model's prediction has remained unchanged or the deviation between the prediction and the adversarial target has increased as a result of the attack. 
However, if $\rmse \left(\hat y_{adv}, y_{adv} \right)$ decreases relative to $\rmse \left(\hat y, y_{adv} \right)$, the \gls{drs} converges to zero and the deformation robustness drops exponentially, see Fig. \ref{fig:drs} in Appendix \ref{secA3:robustness_scores}. \\
Analogously, the \gls{drs} can also be defined for semi-targeted attacks with lower and upper bound constraints:
\begin{equation}
    \drs \left( \hat{y}, \hat{y}_{adv}, a, b \right) = \min \left( \exp \left( 1 - \frac{\brmse _{[a,b]} \left(\hat{y}\right)}{\brmse _{[a,b]} \left(\hat{y}_{adv}\right) + \gamma}\right), 1 \right)
\end{equation}
Here, the \gls{drs} measures the extent to which the deviation between the model's prediction and the area enclosed by the lower and upper bound has decreased as a result of the attack.

\subsubsection{Total adversarial robustness}
\label{subsubsec:total_adversarial_robustness}
Neither the \gls{prs} nor the \gls{drs} individually provide a thorough assessment of a regression model's robustness to targeted or semi-targeted attacks.
While the \gls{prs} only captures the impact of an attack on the model's performance, the \gls{drs} solely measures how the attack affected the deviation between the model's prediction and the attacker's target or the attacker's constraints.
From the victim's perspective, a model is only considered robust if it has both a high \gls{prs} and a high \gls{drs}.
We therefore define the \gls{tars}, which combines the \gls{prs} and the \gls{drs} into one score. Thus, the \gls{tars} provides a comprehensive measure of a model's robustness:
\begin{equation}
    \tars _{\beta} = \left( 1 + \beta ^2 \right) \frac{\prs \cdot \drs}{\left( \beta ^2 \cdot \prs \right) + \drs}
\end{equation}
Note that the \gls{tars} is inspired by the $F_{\beta}$ score and uses a parameter $\beta \in \mathbb{R}^+$. 
In the case $\beta = 1$, the \gls{tars} is the harmonic mean between \gls{drs} and \gls{prs}. 
Depending on the application, $\beta$ can be adjusted such that the \gls{drs} is considered to be $\beta$ times as important as the \gls{prs}.
Thus, for $\beta > 1$, deformation robustness is weighted higher, whereas for $\beta < 1$, performance robustness is given more weight. 
Compared to weighted arithmetic averaging, the \gls{tars} has the advantage that a model's robustness is only considered high if it has both high performance robustness and high deformation robustness. 
However, if either the \gls{prs} or the \gls{drs} is very low, the \gls{tars} also quantifies the robustness of the model as being poor, see Figure \ref{fig:tars} in Appendix \ref{secA3:robustness_scores}.
We recommend calculating the \gls{tars} for all relevant adversarial targets and constraints individually. 
This allows a better assessment of which targets or constraints the model is particularly susceptible to.
Also, a threat analysis \cite{bitton2023evaluating} should be conducted in advance for the use case of interest.
In this way, various important attack scenarios and the associated targets and constraints of an attacker can be identified.

\section{Experimental setup}
\label{sec:experiments}
As contribution (C3), we investigated the robustness of two \gls{dl}-based wind power forecasting models to adversarial attacks. 
Besides a forecasting model for individual wind farms, we also considered a forecasting model for predicting the wind power generation in the whole of Germany. 
Furthermore, as contribution (C4), we examined to what extent adversarial training can increase the robustness of the two models. 
In the following, the experimental setup is described in more detail. 

\subsection{Data}
\label{subsec:data}
To predict the power generation of individual wind farms, we used the wind power measurements and wind speed predictions of the 10 different wind farms from the publicly available GEFCom2014 wind forecasting dataset \cite{hong2016probabilistic}. 
The wind speed predictions were generated for the locations of the wind farms and are univariate time series. 
A separate \gls{lstm} model for wind power forecasting was trained for each of the 10 wind farms.
For training and hyperparameter tuning of the forecasting models, the data of each wind farm were divided into training, validation and test datasets. 
To forecast the wind power generated throughout Germany, real and publicly available wind power data and wind speed forecasts were used as well. 
The wind speed forecasts were aggregated to 100 × 85 weather maps covering Germany.
Using blocked cross-validation, the dataset was divided into 8 different subsets. 
For each of the 8 subsets, a separate \gls{cnn} model was trained to forecast wind power generation across Germany.
To this end, each subset was divided into a training, validation, and test dataset.
The wind power and wind speed data from both the individual wind farm dataset and the Germany dataset had an hourly frequency.
For more information on both datasets, see Appendices \ref{subsecA1:data_wind_farm} and \ref{subsecA1:data_germany}.

\subsection{Forecasting models}
\label{subsec:forecasting_models}
We used an encoder-decoder \gls{lstm} \cite{sutskever2014sequence} for a multi-step ahead forecast of the power generated by individual wind farms, similar to \cite{lu2018short}. 
First, the encoder \gls{lstm} network encoded an input sequence consisting of the wind power measurements for the past 12 hours into a latent representation.
Using the latent representation and wind speed predictions for the forecast horizon, the decoder \gls{lstm} network then sequentially generated a wind power forecast for the next 8 hours with hourly time resolution. \\
To forecast the wind power generated across Germany, we used the approach of \cite{bosma2022estimating}.
Here, a \gls{cnn} model was applied to forecast the wind power based on weather maps. 
We used a ResNet-34 \cite{he2016deep} to make an 8-hour forecast with hourly resolution for the wind energy generated throughout Germany. 
This model was sequentially applied to the wind speed maps. 
It forecasted the wind power generation of a particular point in time based on the wind speed forecasts for the 5 hours leading up to the estimation time.
The two models are described more detailed in Appendices \ref{subsecA2:wp_forecast_single} and \ref{subsecA2:wp_forecast_germany}.

\subsection{Adversarial robustness evaluation}
\label{subsec:adversarial_robustness}
\begin{figure}[h]
  \begin{minipage}[t]{0.33\textwidth}
  \caption{Four different adversarial targets considered for the targeted \gls{pgd} attacks: the prediction of increasing (solid), decreasing (dashed), constant (dotted), and zig-zag shaped (dash-dotted) generated wind power}
    \label{fig:adversarial_targets}
  \end{minipage} \hfill
  \begin{minipage}[t]{0.64\textwidth}
   \vspace{0pt}
   \includegraphics[trim={0 0 0 2cm},clip, width=\textwidth]{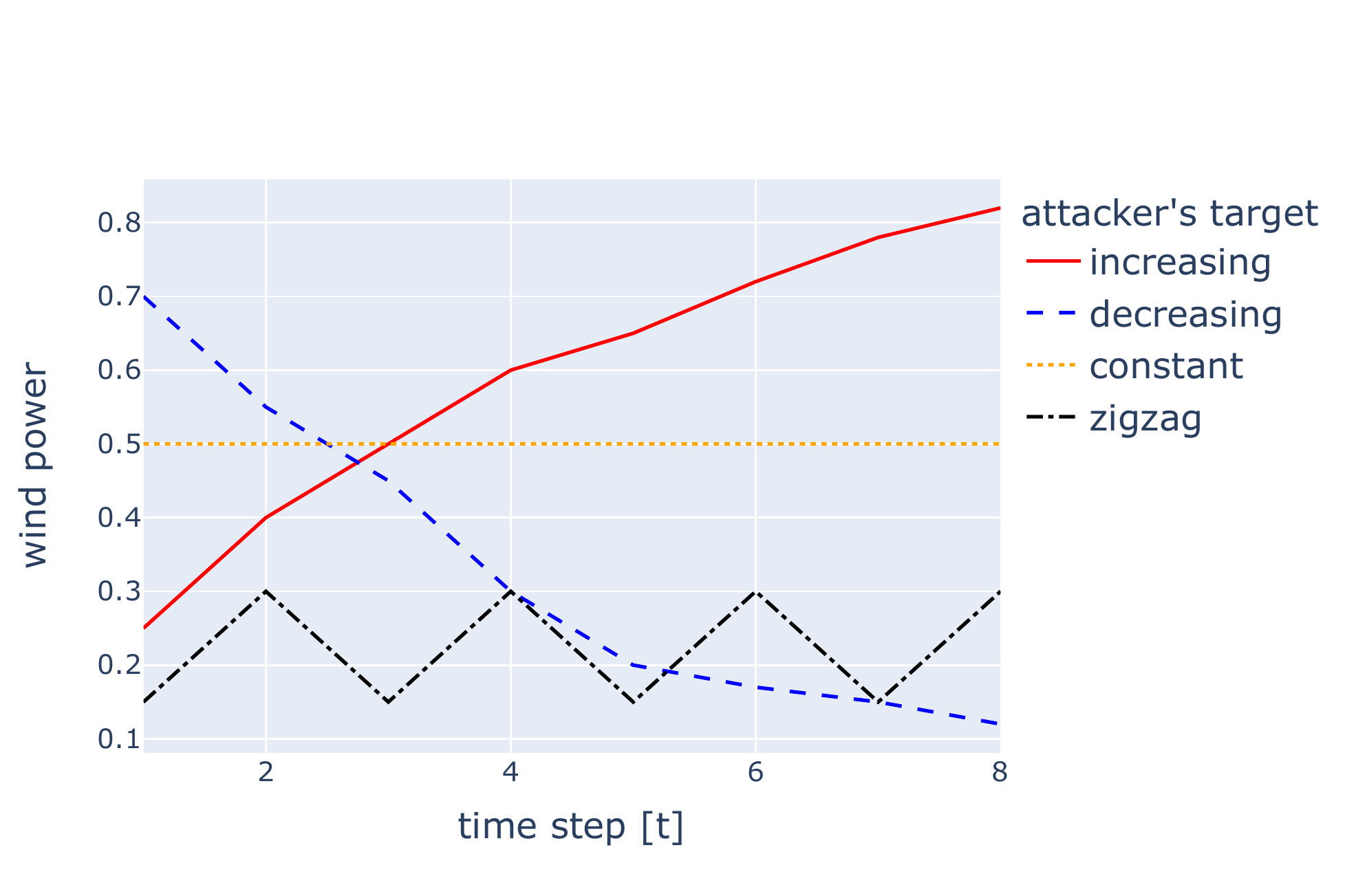}
  \end{minipage}
\end{figure}

\begin{figure}[h]
  \begin{minipage}[t]{0.33\textwidth}
    \caption{Four different constraints considered for the semi-targeted \gls{pgd} attacks: the forecast has to be between 0.75 and 1.0 (horizontal mesh), 0.5 and 0.75 (right diagonal), 0.25 and 0.5 (diagonal mesh), or between 0.0 and 0.25 (left diagonal)}
    \label{fig:adversarial_constraints}
  \end{minipage} \hfill
  \begin{minipage}[t]{0.64\textwidth}
   \vspace{0pt}
   \includegraphics[trim={0 0 0 2cm},clip, width=\textwidth]{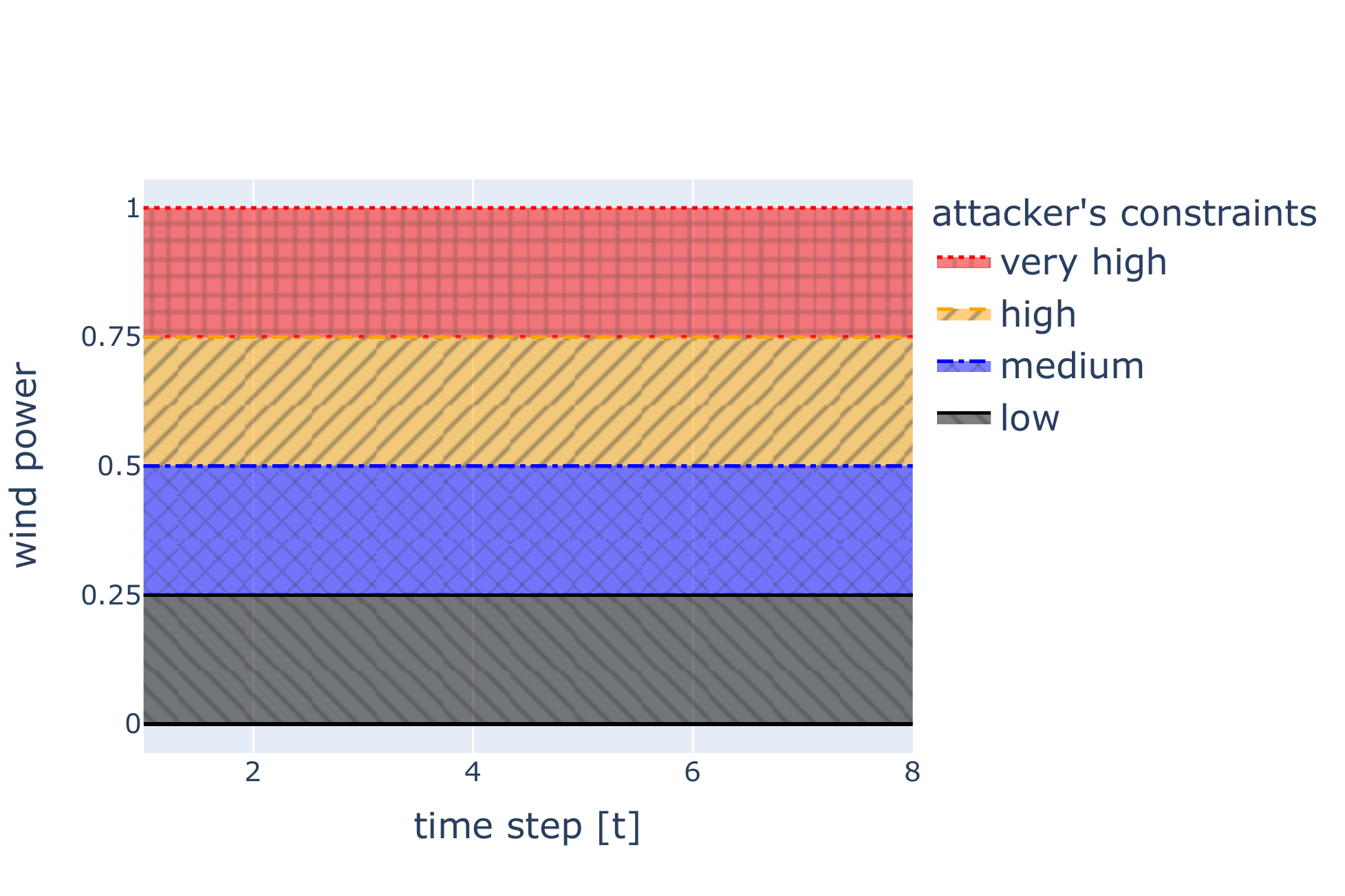}
  \end{minipage}
\end{figure}

We investigated the susceptibility of the two forecasting models to adversarial noise attacks, as well as untargeted, semi-targeted, and targeted \gls{pgd} attacks.
In all attacks, only the standardized wind speeds were manipulated.
We considered perturbations with a maximum magnitude of $\epsilon = 0.15$ within the $L_{\infty}$ norm ball.
Here, $\epsilon$ was chosen such that the maximum possible perturbation corresponds to a change in wind speed of about 0.5 m/s.
According to the maximum derivative of a reference wind turbine's power curve, these perturbations should never cause a change in the generated wind power of more than 10\% of the rated power.
The reference wind turbine was an Enercon E-115\footnote{The Enercon E-115 was chosen as the reference wind turbine because in 2016, 2017, and 2018, Enercon was the market-leading manufacturer in Germany and its most installed turbine type in each of these years was the E-115, according to \cite{unnewehr2021getting}.}. \\
In the experiments, we examined repeated noise attacks with Gaussian noise and 100 repetitions.
For the \gls{pgd} attacks, we used $T = 100$ \gls{pgd} steps\footnote{The number of steps $T$ was chosen such that doubling $T$ does not increase the success rate of the attack, as proposed by \cite{carlini2019evaluating}.} with a step size\footnote{This choice of the step size ensures that the maximum perturbation magnitude $\epsilon$ can be reached with the number of steps $T$.} of $\alpha = 2 \epsilon / T$. 
The targeted attacks were generated for a total of 4 different adversarial targets, as shown in Fig. \ref{fig:adversarial_targets}.
Among these, 3 targets correspond to various realistic scenarios. 
They aim to manipulate the model such that either increasing, decreasing, or constant wind power is predicted. 
In contrast, the fourth scenario corresponds to a zigzag line. 
This target was used to investigate how arbitrarily the forecasts can be manipulated.
In addition, semi-targeted attacks were generated for a total of 4 different lower and upper bound constraints, as shown in Fig. \ref{fig:adversarial_constraints}.
The objective of these constraints is to manipulate the model's predictions so that the forecasted wind power is either in a low, medium, high, or very high range.
Furthermore, we investigated to what extent the adversarial robustness of the two models can be increased with the help of adversarial training. 
For this purpose, adversarial examples were generated in each training iteration by perturbing every training sample using the untargeted \gls{pgd} attack.
The above described parameters were used here for the the untargeted \gls{pgd} attack as well.
The model was then trained on the adversarial examples only. \\
While the robustness of the two models to untargeted attacks was assessed using only the \gls{prs}, the robustness to semi-targeted and targeted attacks was quantified using all three scores (\gls{prs}, \gls{drs}, and \gls{tars}). They were calculated individually for each target and constraint of the attacker.
This was done by first generating an adversarial example from every test sample.
Then, the \gls{prs}, \gls{drs} and \gls{tars} were calculated sample-wise.
Next, the average \gls{prs}, \gls{drs}, and \gls{tars} were calculated for each individual test dataset by averaging the scores of the respective test samples.
Finally, the means and standard deviations of the average \gls{prs}, \gls{drs} and \gls{tars} were calculated from the 10 individual wind farm test datasets and the 8 Germany test datasets, respectively.

\section{Results}
\label{sec:results}

\subsection{Adversarial robustness of the LSTM model}
\label{subsec:robustness_single}
The forecasting model for wind farms was quite robust to untargeted adversarial attacks with $\epsilon = 0.15$, as Table \ref{tab:pgd_untargeted_single-turbine} shows.
While the ordinarily trained model achieved an average \gls{rmse} of 12.90\% of installed capacity\footnote{In wind power forecasting, it is common to express the \gls{rmse} as a percentage of installed capacity. To obtain the percentage value, we multiply the \gls{rmse} calculated from Equation \ref{eq:rmse} by 100, as all wind power measurements in our work are normalized by installed capacity.} when not under attack, its performance deteriorated to an average \gls{rmse} of 15.30\% when attacked by untargeted \gls{pgd} attacks.
The \gls{prs} was thus 0.79 in the case of untargeted \gls{pgd} attacks. 
Noise attacks had an even lower impact on the prediction quality of the model and achieved an average \gls{prs} value of 0.96. \\
Semi-targeted \gls{pgd} attacks had the highest impact when the constraint required the prediction of medium wind power, as shown in Table \ref{tab:pgd_semitargeted_single-turbine}.
For this constraint, an average \gls{tars} of 0.78 was obtained for the ordinarily trained model. For the other three constraints, the average \gls{tars} was 0.79 or more. 
Thus, the model was robust to semi-targeted \gls{pgd} attacks as well. \\
As shown in Table \ref{tab:pgd_targeted_single-turbine}, targeted \gls{pgd} attacks with $\epsilon = 0.15$ had a similar impact on the \gls{lstm} forecasting model for all four adversarial targets.
Here, the ordinarily trained model achieved an average \gls{tars} value of 0.86 or greater for each of the attacker's targets. It was thus very robust to this type of attack.
\begin{table}[h]
\begin{center}
\caption{Mean \gls{prs} and \gls{rmse} values with standard deviation for the \gls{lstm} forecasting model when attacked by noise attacks and untargeted \gls{pgd} attacks}
\label{tab:pgd_untargeted_single-turbine}
    \begin{tabular}{@{}lccccc@{}}
    \toprule
         \multicolumn{1}{@{}c@{}}{} &
         \multicolumn{2}{@{}c@{}}{\textbf{ordinary training}} & &
         \multicolumn{2}{@{}c@{}}{\textbf{adversarial training}}
         \\
         \cmidrule(lr){2-3} \cmidrule(lr){5-6}
        \textbf{attack} & \gls{prs}  & \gls{rmse} [\%] & & \gls{prs}  & \gls{rmse} [\%] \\
    \midrule 
        No attack & - & 12.90 $\pm$ 1.21 && - & 13.24 $\pm$ 1.22\\
        \gls{pgd} & 0.79 $\pm$ 0.02 & 15.30 $\pm$ 1.29 && 0.84 $\pm$ 0.02 & 14.99 $\pm$ 1.31 \\
        Noise & 0.96 $\pm$ 0.01 & 13.01 $\pm$ 1.19 && 0.98 $\pm$ 0.00 & 13.29 $\pm$ 1.22 \\
    \bottomrule
\end{tabular}	 		
\end{center}
\end{table}

\begin{table}[h]
\begin{center}
\caption{Mean \gls{tars}, \gls{drs}, and \gls{prs} values with standard deviation for the \gls{lstm} forecasting model under semi-targeted \gls{pgd} attacks}
\label{tab:pgd_semitargeted_single-turbine}%
    \begin{tabular}{@{}lcccccc@{}}
    \toprule
         \multicolumn{1}{@{}l@{}}{\textbf{attacker's}} &
         \multicolumn{3}{@{}c@{}}{\textbf{ordinary training}} &
         \multicolumn{3}{@{}c@{}}{\textbf{adversarial training}}
         \\
         \cmidrule(lr){2-4} \cmidrule(lr){5-7}
        \textbf{constraints} & \gls{tars} & \gls{drs}  & \gls{prs}  & \gls{tars} & \gls{drs}  & \gls{prs} \\
    \midrule
        low & 0.79 $\pm$ 0.02 & 0.81 $\pm$ 0.02 & 0.86 $\pm$ 0.02& 0.84 $\pm$ 0.02 & 0.84  $\pm$ 0.02 & 0.90 $\pm$ 0.02\\
        medium & 0.78 $\pm$ 0.02 & 0.78 $\pm$ 0.02 & 0.86 $\pm$ 0.02 & 0.82 $\pm$ 0.02 & 0.82 $\pm$ 0.02 & 0.89 $\pm$ 0.02 \\
        high & 0.80 $\pm$ 0.02 & 0.82 $\pm$ 0.03 & 0.86 $\pm$ 0.02 & 0.85 $\pm$ 0.03 & 0.85 $\pm$ 0.03 & 0.90 $\pm$ 0.02 \\
        very high & 0.84 $\pm$ 0.02 & 0.87 $\pm$ 0.02 & 0.87 $\pm$ 0.02 & 0.88 $\pm$ 0.02 & 0.90 $\pm$ 0.02 & 0.91 $\pm$ 0.02 \\
    \bottomrule
\end{tabular}   	
\end{center}
\end{table}

\begin{figure}[h]
\centering
\begin{subfigure}[t]{\textwidth}
  \centering
    \includegraphics[ width=\textwidth]{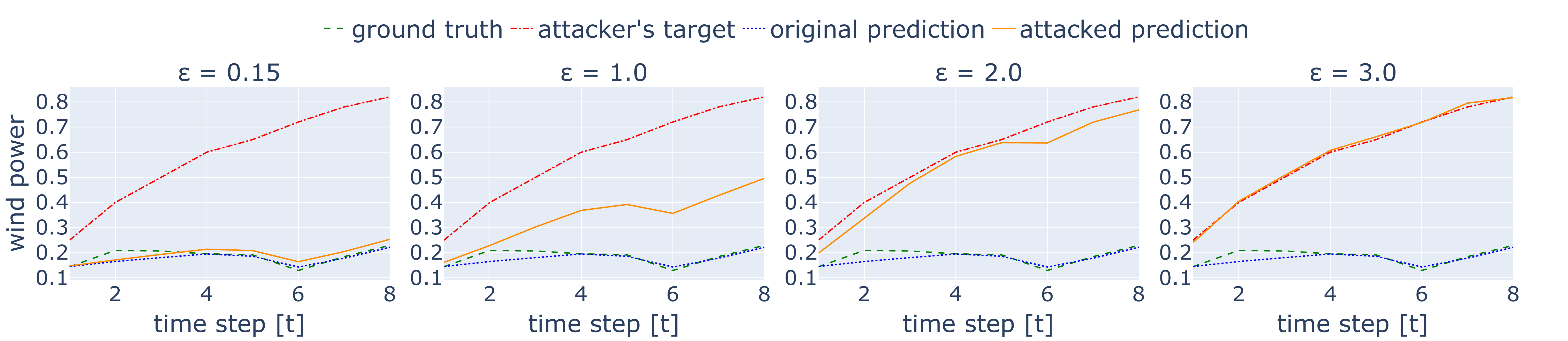}
     \caption{While the original prediction (dotted) approximates the ground truth (dashed) very well, the attacked prediction (solid) converges to the attacker's target (dash-dotted) with increasing maximum perturbation magnitude $\epsilon$}
  \label{fig:st_predictions}
\end{subfigure}
\begin{subfigure}[t]{\textwidth}
  \centering
    \includegraphics[width=\textwidth]{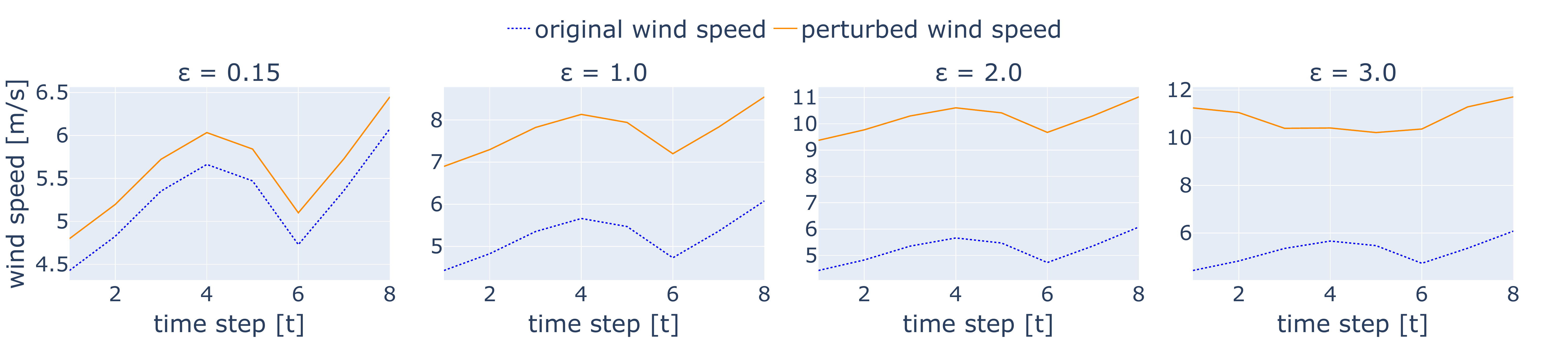}
    \caption{As the perturbation magnitude $\epsilon$ rises, the perturbed wind speeds (solid) increasingly diverge from the original wind speeds (dotted). In addition, the shapes of the perturbed wind speeds often resemble the shapes of the attacked predictions in (a)}
  \label{fig:st_perturbations}
\end{subfigure}
\caption{Four targeted \gls{pgd} attacks with maximum perturbation magnitudes $\epsilon=0.15$ (left), $\epsilon=1.0$ (center-left), $\epsilon=2.0$ (center-right), and $\epsilon=3.0$ (right) on an exemplary prediction of the \gls{lstm} forecasting model.
The figures show the impact of the attacks on (a) the wind power forecast and (b) the input data}
\label{fig:single-turbine_example-attack}
\end{figure}

\begin{table}[h]
\begin{center}
\caption{Mean \gls{tars}, \gls{drs}, and \gls{prs} values with standard deviation for the \gls{lstm} forecasting model when attacked by targeted \gls{pgd} attacks}
\label{tab:pgd_targeted_single-turbine}%
    \begin{tabular}{@{}lcccccc@{}}
    \toprule
         \multicolumn{1}{@{}l@{}}{\textbf{attacker's}} &
         \multicolumn{3}{@{}c@{}}{\textbf{ordinary training}} &
         \multicolumn{3}{@{}c@{}}{\textbf{adversarial training}}
         \\
         \cmidrule(lr){2-4} \cmidrule(lr){5-7}
        \textbf{target} & \gls{tars} & \gls{drs}  & \gls{prs}  & \gls{tars} & \gls{drs}  & \gls{prs} \\
        \midrule
        increasing & 0.89 $\pm$ 0.01 & 0.91 $\pm$ 0.01 & 0.88 $\pm$ 0.01 & 0.92 $\pm$ 0.01 & 0.94 $\pm$ 0.01 & 0.91 $\pm$ 0.02 \\
        decreasing & 0.90 $\pm$ 0.01 & 0.91 $\pm$ 0.01 & 0.90 $\pm$ 0.01 & 0.93 $\pm$ 0.01 & 0.94 $\pm$ 0.01 & 0.93 $\pm$ 0.02 \\
        constant & 0.86 $\pm$ 0.02 & 0.87 $\pm$ 0.02 & 0.88 $\pm$ 0.01 & 0.90 $\pm$ 0.02 & 0.90 $\pm$ 0.02 & 0.91 $\pm$ 0.02 \\
        zigzag & 0.89 $\pm$ 0.01 & 0.89 $\pm$ 0.01 & 0.89 $\pm$ 0.01 & 0.92 $\pm$ 0.01 & 0.92 $\pm$ 0.01 & 0.93 $\pm$ 0.02 \\
    \bottomrule
\end{tabular}
\end{center}
\end{table}
\noindent
In order to achieve successful targeted \gls{pgd} attacks on the ordinarily trained forecasting model, very strong perturbations of the wind speed time series were required, as the example in Figure \ref{fig:single-turbine_example-attack} shows.
Here, the attacked prediction did not closely match the attacker's target until the perturbation magnitude was $\epsilon = 3.0$.
In addition, the perturbed wind speed time series often had a shape similar to the shape of the wind power forecast.
This indicates that the model's behavior was physically correct.
\newline
With the help of adversarial training, the model's robustness to \gls{pgd} attacks and noise attacks could be slightly increased, as shown by the respective \gls{prs} values in Table \ref{tab:pgd_untargeted_single-turbine} along with the \gls{tars} values in Table \ref{tab:pgd_semitargeted_single-turbine} and Table \ref{tab:pgd_targeted_single-turbine}. 
However, when not under attack, the forecast accuracy of the model slightly deteriorated due to adversarial training. 
Thus, the average \gls{rmse} value between the model's predictions and the ground truth on the test datasets was about 12.90\% of installed capacity in the case of ordinary training, but 13.24\% in the case of adversarial training.

\subsection{Adversarial robustness of the CNN model}
\label{subsec:robustness_germany}
In contrast to the \gls{lstm} forecasting model for the wind farms, the \gls{cnn} model for forecasting the wind power generation throughout Germany was very susceptible to \gls{pgd} attacks with $\epsilon = 0.15$. 
The average \gls{prs} value for untargeted \gls{pgd} attacks on the ordinarily trained model was 0.05, as shown in Table \ref{tab:pgd_untargeted_germany}.
As a result of the untargeted \gls{pgd} attacks, the average \gls{rmse} of the model deteriorated from 5.24\% of installed capacity to 46.18\%.
Noise attacks resulted in an average \gls{prs} of 0.93 for the ordinarily trained model. 
Thus, they had a similarly small impact on the \gls{cnn} forecasting model as on the \gls{lstm} forecasting model. \\
The ordinarily trained \gls{cnn} model was also very vulnerable to semi-targeted and targeted \gls{pgd} attacks. 
For the semi-targeted attacks, the \gls{tars} for all four constraints was 0.10 or less, as shown in Table \ref{tab:pgd_semitargeted_germany}. 
As Table \ref{tab:pgd_targeted_germany} shows, the average \gls{tars} value for the targeted attacks with the increasing target was 0.01. 
For the zigzag shaped as well as the constant and decreasing target of the attacker, the average \gls{tars} was even 0.00.
\begin{table}[h]
\begin{center}
\caption{Mean \gls{prs} and \gls{rmse} values with standard deviation for the \gls{cnn} forecasting model when attacked by noise attacks and untargeted \gls{pgd} attacks}
\label{tab:pgd_untargeted_germany}%
    \begin{tabular}{@{}lccccc@{}}
    \toprule
         \multicolumn{1}{@{}c@{}}{} &
         \multicolumn{2}{@{}c@{}}{\textbf{ordinary training}} & &
         \multicolumn{2}{@{}c@{}}{\textbf{adversarial training}}
         \\
         \cmidrule(lr){2-3} \cmidrule(lr){5-6}
        \textbf{attack} & \gls{prs}  & \gls{rmse} [\%] & & \gls{prs}  & \gls{rmse} [\%] \\
        \midrule
        No attack & - & 5.24 $\pm$ 1.17 && - & 6.22 $\pm$ 1.53 \\
        \gls{pgd} & 0.05 $\pm$ 0.04 & 46.18 $\pm$ 10.93 && 0.82 $\pm$ 0.08 & 7.77 $\pm$ 2.08 \\
        Noise & 0.93 $\pm$ 0.04 & 5.22 $\pm$ 1.10 && 0.99 $\pm$ 0.00 & 6.21 $\pm$ 1.55 \\
    \bottomrule
\end{tabular}	
\end{center}
\end{table}

\begin{table}[h]
\begin{center}
\caption{Mean \gls{tars}, \gls{drs}, and \gls{prs} values with standard deviation for the \gls{cnn} forecasting model under semi-targeted \gls{pgd} attacks}
\label{tab:pgd_semitargeted_germany}%
    \begin{tabular}{@{}lcccccc@{}}
    \toprule
         \multicolumn{1}{@{}l@{}}{\textbf{attacker's}} &
         \multicolumn{3}{@{}c@{}}{\textbf{ordinary training}} &
         \multicolumn{3}{@{}c@{}}{\textbf{adversarial training}}
         \\
         \cmidrule(lr){2-4} \cmidrule(lr){5-7}
        \textbf{constraints} & \gls{tars} & \gls{drs}  & \gls{prs}  & \gls{tars} & \gls{drs}  & \gls{prs} \\
        \midrule
        low & 0.10 $\pm$ 0.05 & 0.59 $\pm$ 0.15 & 0.12 $\pm$ 0.07 & 0.67 $\pm$ 0.11 & 0.71 $\pm$ 0.09 & 0.83 $\pm$ 0.07 \\
        medium & 0.06 $\pm$ 0.04 & 0.18 $\pm$ 0.09 & 0.09 $\pm$ 0.06 & 0.46 $\pm$ 0.07 & 0.54 $\pm$ 0.07 & 0.64 $\pm$ 0.10 \\
        high & 0.01 $\pm$ 0.02 & 0.04 $\pm$ 0.06 & 0.06 $\pm$ 0.06 & 0.61 $\pm$ 0.08 & 0.78 $\pm$ 0.07 & 0.62 $\pm$ 0.12 \\
        very high & 0.01 $\pm$ 0.03 & 0.06 $\pm$ 0.09 & 0.03 $\pm$ 0.06 & 0.66 $\pm$ 0.09 & 0.89 $\pm$ 0.06 & 0.61 $\pm$ 0.11 \\
    \bottomrule
\end{tabular} 	 	
\end{center}
\end{table}

\begin{figure}[H]
\centering
\begin{subfigure}[t]{\textwidth}
    \begin{minipage}[t]{0.33\textwidth}
  \caption{While the original prediction (dotted) matches the ground truth (dashed) very well, the attacked prediction (solid) is much closer to the attacker's target (dash-dotted) than to the ground truth}
  \label{subfig:germany_example-attack_predictions}
  \end{minipage} \hfill
  \begin{minipage}[t]{0.64\textwidth}
   \vspace{0pt}
   \includegraphics[trim={0 0 0 2cm},clip, width=\textwidth]{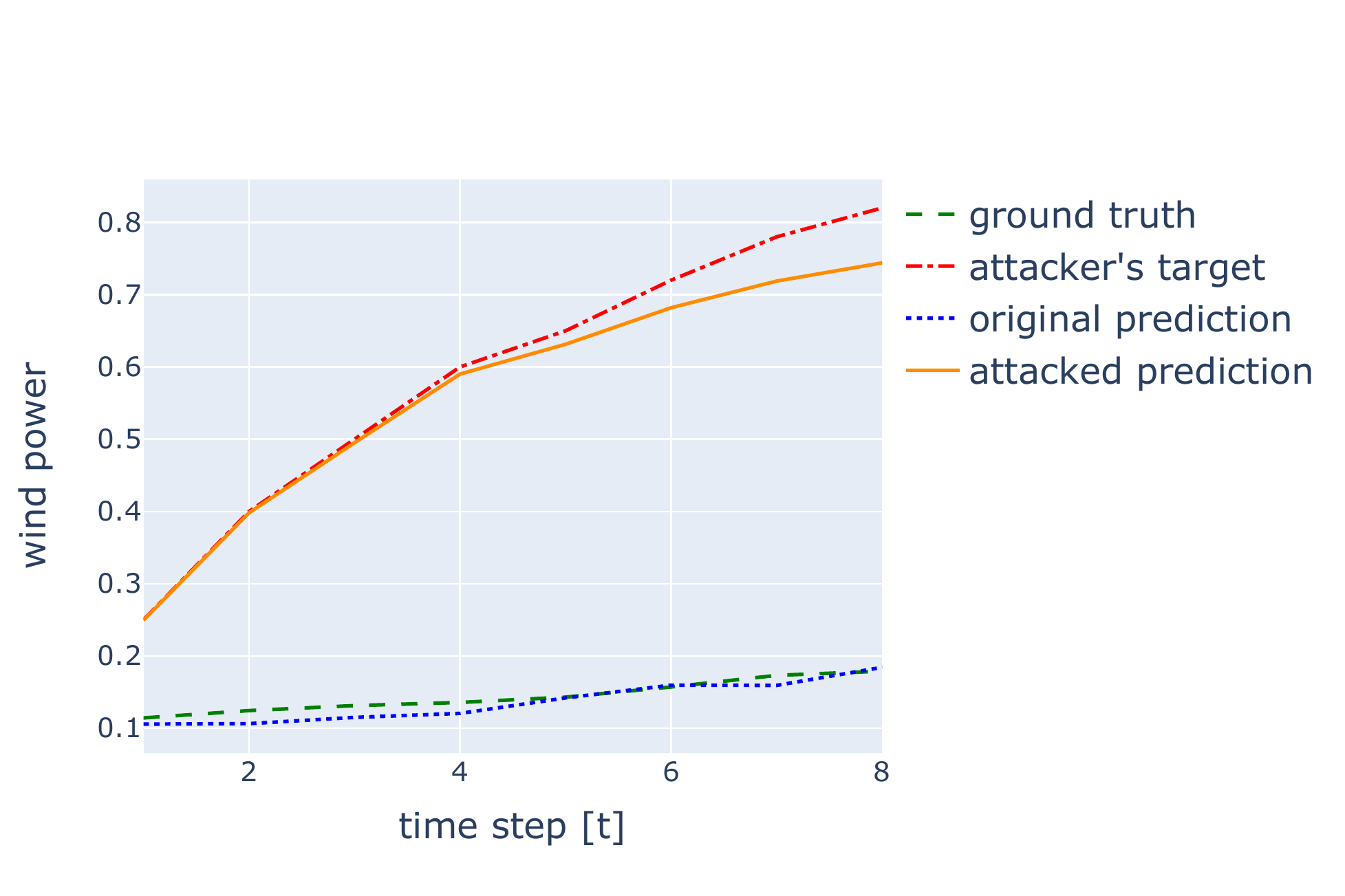}
  \end{minipage}
\end{subfigure}
\begin{subfigure}[t]{\textwidth}
  \centering
    \includegraphics[width=\textwidth]{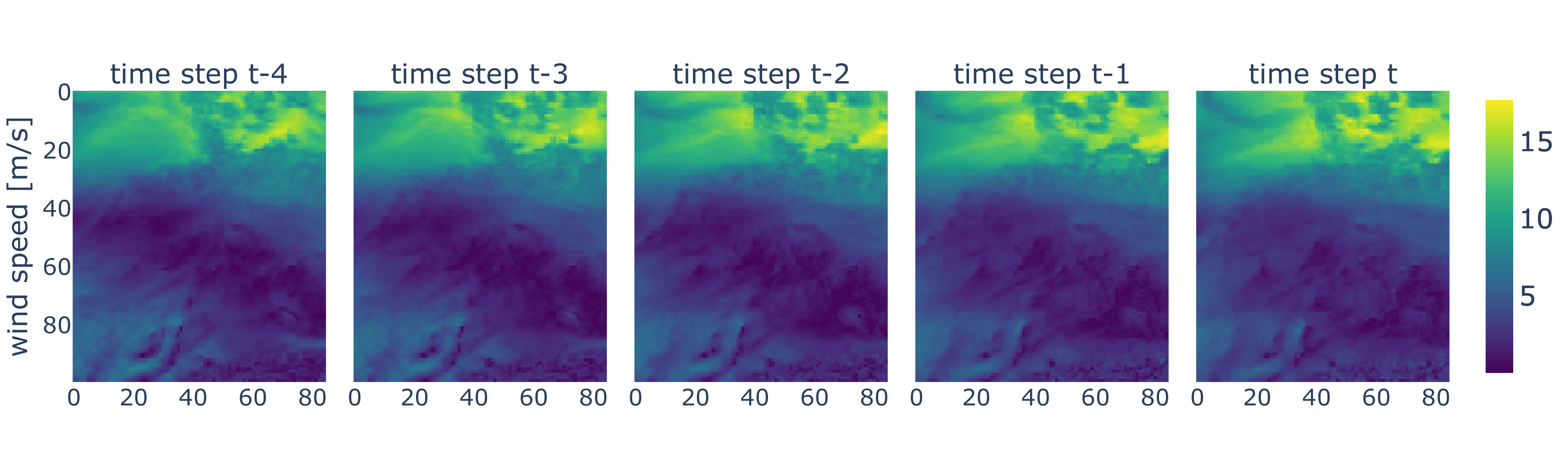}
    \caption{The original wind speeds used to predict the last time step of the forecast ($t=8$)}
  \label{subfig:germany_example-attack_input_original}
\end{subfigure}
\begin{subfigure}[t]{\textwidth}
  \centering
    \includegraphics[width=\textwidth]{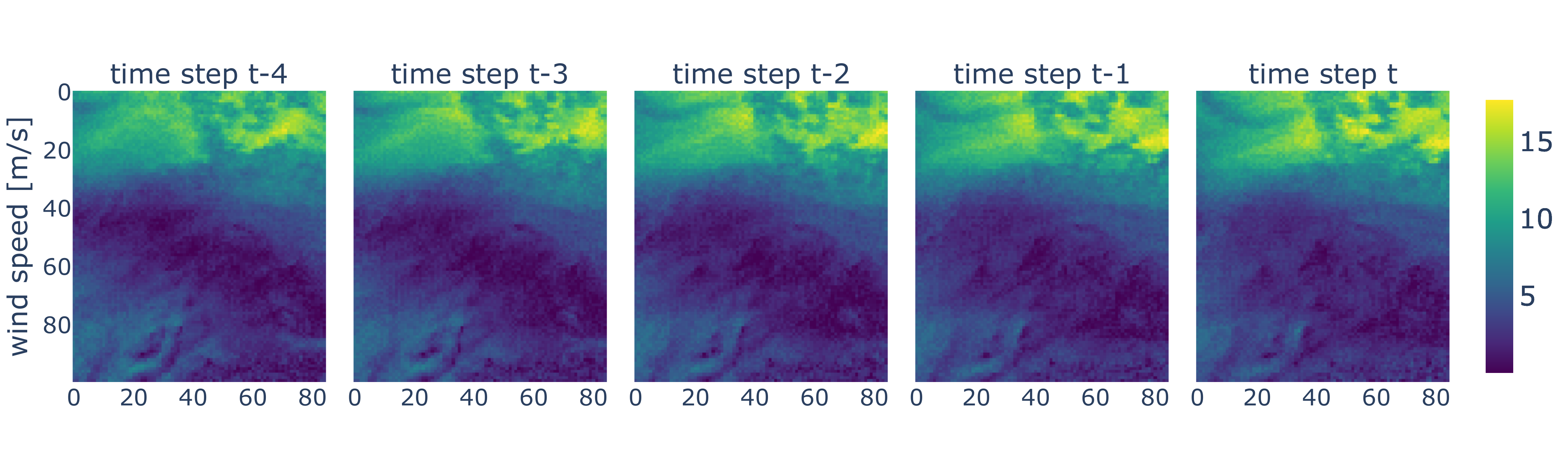}
    \caption{The wind speeds from (b) with the perturbations caused by the \gls{pgd} attack}
  \label{subfig:germany_example-attack_input_perturbed}
\end{subfigure}
\begin{subfigure}[t]{\textwidth}
  \centering
    \includegraphics[width=\textwidth]{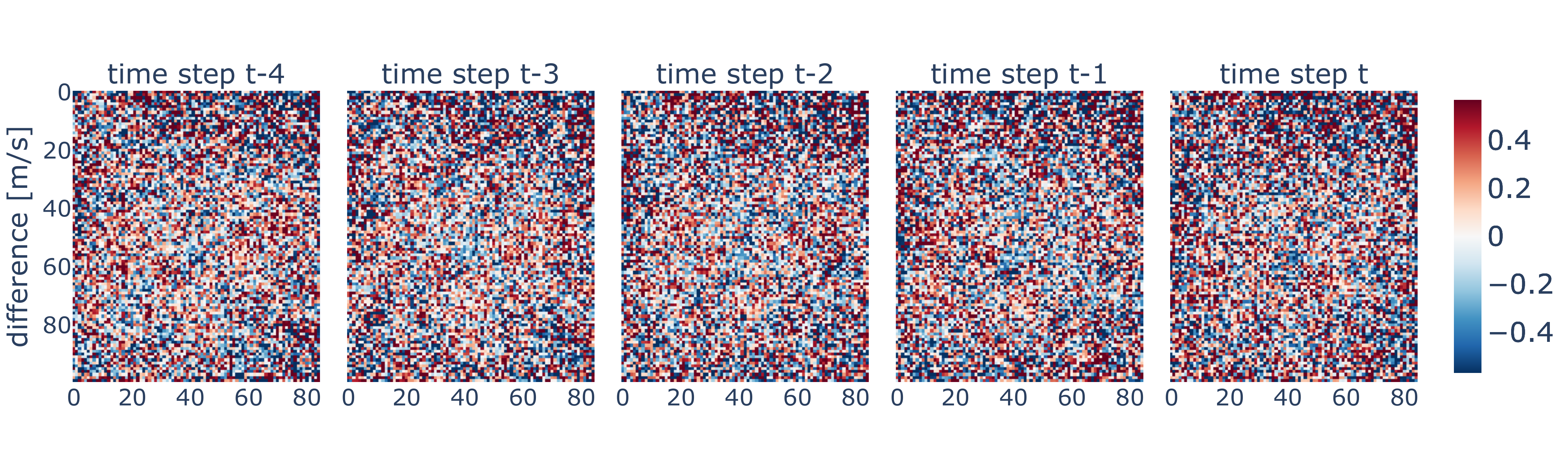}
    \caption{Difference between the perturbed (c) and original (b) wind speeds, i.e., $x_{adv} - x$}
  \label{subfig:germany_example-attack_input_difference}
\end{subfigure}
\caption{A targeted \gls{pgd} attack with perturbation magnitude $\epsilon=0.15$ on an exemplary prediction of the \gls{cnn} forecasting model.
The figures show (a) the impact of the attack on the wind power forecast as well as (b) the original input data, (c) the perturbed input data, and (d) the difference between the original and perturbed input data for the last time step of the forecast. All weather maps shown represent wind speeds across Germany in the unit m/s}
\label{fig:germany_example-attack}
\end{figure}

\begin{figure}[H]
    \centering
    \includegraphics[trim={0cm 0cm 0cm 2cm}, clip, width=\textwidth]{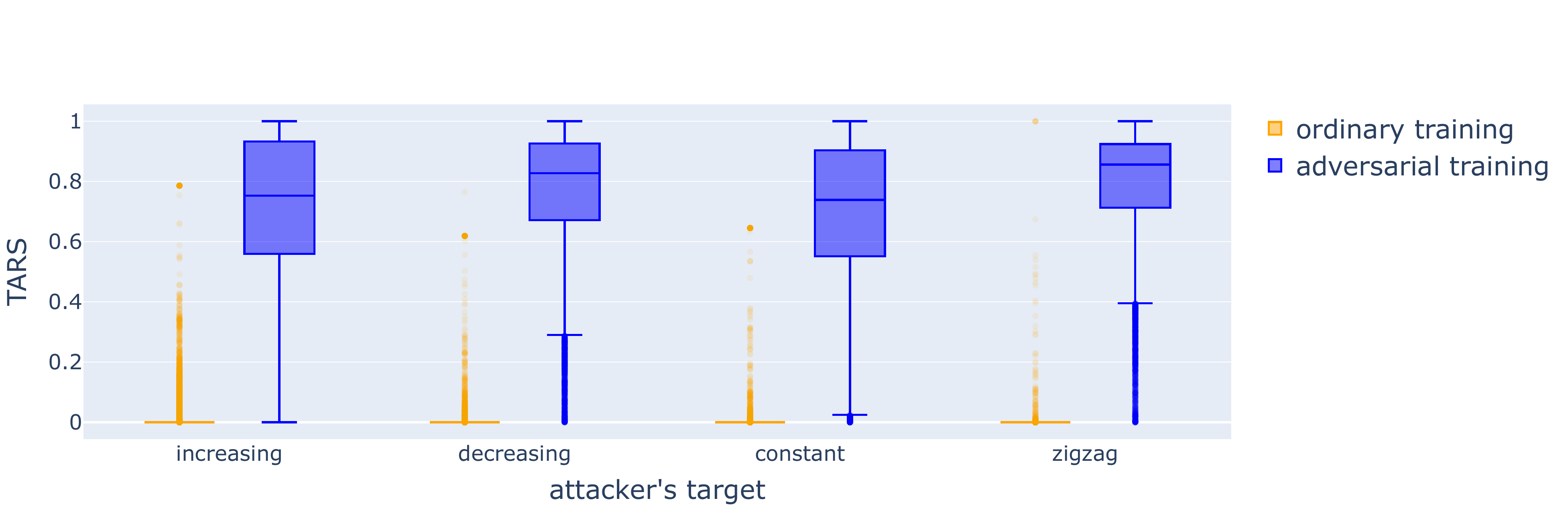}
    \caption{
    \gls{tars} values of targeted \gls{pgd} attacks on the \gls{cnn} forecasting model for the
    increasing (left), decreasing (center-left), constant (center-right), and zigzag (right) target. The boxplots show that in the case of ordinary training (orange) the attacks are successful for most test samples.
    If adversarial training (blue) is used instead, the effects of the attacks are significantly reduced
    }
    \label{fig:boxplot_targeted_attacks_tars_germany}
\end{figure}

\begin{table}[H]
\begin{center}
\caption{Mean \gls{tars}, \gls{drs}, and \gls{prs} values with standard deviation for the \gls{cnn} forecasting model when attacked by targeted \gls{pgd} attacks}
\label{tab:pgd_targeted_germany}%
    \begin{tabular}{@{}lcccccc@{}}
    \toprule
         \multicolumn{1}{@{}l@{}}{\textbf{attacker's}} &
         \multicolumn{3}{@{}c@{}}{\textbf{ordinary training}} &
         \multicolumn{3}{@{}c@{}}{\textbf{adversarial training}}
         \\
         \cmidrule(lr){2-4} \cmidrule(lr){5-7}
        \textbf{target} & \gls{tars} & \gls{drs}  & \gls{prs}  & \gls{tars} & \gls{drs}  & \gls{prs} \\
        \midrule
        increasing & 0.01 $\pm$ 0.03 & 0.04 $\pm$ 0.07 & 0.07 $\pm$ 0.06 & 0.71 $\pm$ 0.08 & 0.90 $\pm$ 0.03 & 0.65 $\pm$ 0.11 \\
        decreasing & 0.00 $\pm$ 0.01 & 0.01 $\pm$ 0.02 & 0.08 $\pm$ 0.05 & 0.77 $\pm$ 0.06 & 0.88 $\pm$ 0.03 & 0.73 $\pm$ 0.08 \\
        constant & 0.00 $\pm$ 0.01 & 0.01 $\pm$ 0.02 & 0.15 $\pm$ 0.10 & 0.69 $\pm$ 0.08 & 0.85 $\pm$ 0.04 & 0.66 $\pm$ 0.12 \\
        zigzag & 0.00 $\pm$ 0.00 & 0.00 $\pm$ 0.00 & 0.22 $\pm$ 0.08 & 0.79 $\pm$ 0.05 & 0.85 $\pm$ 0.05 & 0.79 $\pm$ 0.05 \\
    \bottomrule
\end{tabular}
\end{center}
\end{table}
\noindent
As an example, Figure \ref{fig:germany_example-attack} shows the impact of a \gls{pgd} attack with the increasing adversarial target on an exemplary prediction. 
In this case, small perturbations of the weather maps had caused the model's prediction to move close to the attacker's target. As a result of the \gls{pgd} attack, the wind speeds of the weather maps are both increased and decreased to varying degrees. Yet, the maximum perturbation magnitude is always less than 0.5 m/s. Although the differences between the perturbed weather maps and the original weather maps are visible, they are mostly inconspicuous. \\
The robustness of the \gls{cnn} model to \gls{pgd} attacks could be significantly increased with the help of adversarial training. 
For instance, the average \gls{prs} for the untargeted \gls{pgd} attacks was 0.82 when adversarial training was used, see Table \ref{tab:pgd_untargeted_germany}.
For semi-targeted and targeted attacks, adversarial training resulted in the average \gls{tars} being above 0.46 for all the attacker's constraints and above 0.69 for all the attacker's targets, see Tables \ref{tab:pgd_semitargeted_germany} and \ref{tab:pgd_targeted_germany}, respectively. \\
As shown in Figure \ref{fig:boxplot_targeted_attacks_tars_germany}, adversarial training had a positive effect on the robustness of the model not only on average, but indeed for most test samples. 
Thus, in the case of targeted \gls{pgd} attacks, the 75th percentile of the \gls{tars} was below $4.49 \cdot 10^{-7}$ for all four of the attacker's targets when the model was trained ordinarily.  
When adversarial training was used instead, the 25th percentile of the \gls{tars} was above 0.55 for all four targets of the attacker.
Although adversarial training significantly increased the robustness of the model, there still were individual samples for which the targeted \gls{pgd} attacks were successful.
In addition, adversarial training had a negative effect on the prediction accuracy of the model when not under attack. 
The average \gls{rmse} value between the model's predictions and the ground truth was 5.24\% of installed capacity on the test datasets for ordinary training, but 6.22\% for adversarial training. 

\section{Discussion}
\label{sec:discussion}
In this work, we investigated the adversarial robustness of two different wind power forecasting models. 
We developed the \gls{tars} to quantify the robustness of the models to targeted and semi-targeted adversarial attacks. 
Our results show that wind power forecasting models which make forecasts for individual wind farms are robust even to powerful adversarial attacks. 
It requires very strong perturbations of the input data to bias the model's predictions toward the attacker's target. 
However, these perturbations are such that they appear to fit the model's predictions from a physical point of view.
Thus, we hypothesize that the model behaves physically correct even in the case of attack. \\
On the other hand, wind power forecasting models, which use weather maps to produce forecasts for entire regions, are very vulnerable to adversarial attacks. 
Even small and barely perceptible perturbations of the input data are sufficient to falsify the forecasts almost arbitrarily. 
We suspect that this is due to the high dimensionality of the input data. 
Forecasting models for individual wind farms process very low-dimensional input data with only a few relevant features. 
In contrast, weather maps represent high-dimensional data with many features being relevant for large-scale wind power forecasting. 
This assumption is consistent with the study of \cite{chattopadhyay2019curse}, which showed that the generation of adversarial attacks benefits from higher dimensionality of input data in the classification setting.
Note that the dimensionality of the input data we used is still comparatively low. In real applications, such as in \cite{bosma2022estimating}, various other weather predictions are used besides wind speed forecasts, e.g., predictions for air pressure, air temperature, and air humidity. 
Such input data gives attackers even more attack possibilities. \\
We also studied adversarial training in order to protect the models from attacks. 
While adversarial training exorbitantly increased the robustness of the \gls{cnn} forecasting model, it had only marginal effects on the robustness of the \gls{lstm} forecasting model. 
Adversarial training also slightly deteriorated the forecast accuracy of both models when not under attack. 
This finding is consistent with several studies in the classification setting \cite{tsipras2018robustness,raghunathan2019adversarial,zhang2019theoretically}, which state that there is a trade-off between robustness and accuracy.
Therefore, an important direction for future work is to develop adversarial defenses that do not negatively impact the performance of forecasting models.
An alternative approach could be to scale several robust wind power forecasts for individual wind farms up to a region, as outlined in \cite{jung2014current}. 
However, it remains to be examined whether such an upscaling approach for regional forecasts is as accurate as forecasts generated from weather maps. 
Another important direction for future work is to extend our method used to generate targeted attacks on forecasting models. 
Currently, we select the various adversarial targets very carefully by hand. 
However, it would be desirable to have techniques for automatically generating realistic, application-specific adversarial targets. 
Such techniques would allow a more comprehensive robustness evaluation.

\section{Conclusion}
\label{sec:conclusion}
In this study, we have shown that the use of \glsfirst{dl} for wind power forecasting can pose a security risk. 
In general, our results are relevant for forecasting in power systems, including solar power and load flow forecasting, among others. 
Adversarial attacks also pose a threat to forecasting models used in other critical infrastructures, for example, the financial and insurance sectors. 
\gls{dl}-based forecasting models which obtain input data from safety-critical interfaces should therefore always be tested for their vulnerability to adversarial attacks before being deployed. 
In order to appropriately quantify the robustness of such models, we proposed the \glsfirst{tars}. 
In case of high vulnerability, adequate defense mechanisms, such as adversarial training, should be used to protect the models from attacks. Finally, our work represents a first study of targeted adversarial attacks for \gls{dl}-based regression models, and we expect this to be a promising area for future research.

\section*{Acknowledgments}
This work was carried out as part of the SecDER project (Fkz. 03EI4028B) funded by the German Federal Ministry for Economic Affairs and Climate Action (BMWK). 
It was also supported by the Competence Center for Cognitive Energy Systems of the Fraunhofer Institute for Energy Economics and Energy System Technology (IEE). 
The establishment of the Competence Center for Cognitive Energy Systems is funded by the Hessian State Government. 
Furthermore, parts of the work were performed within the project RL4CES (Fkz. 01$\vert$S22063), which is funded by the German Federal Ministry of Education and Research (BMBF).
We would like to express our special thanks to the German Weather Service (DWD) for providing the weather data.

\section*{Code availability}
The code for the experiments in this paper, as well as the associated visualizations, is available at \url{https://github.com/FraunhoferIEE/taaowpf}.

\bibliographystyle{unsrt}  
\bibliography{taaowpf_arxiv} 

\appendix

\counterwithout{figure}{section}
\setcounter{figure}{9}

\section{Adversarial robustness scores}
\label{secA3:robustness_scores}
\begin{figure}[H]
  \begin{minipage}[t]{0.32\textwidth}
  \caption{Evolution of the \gls{prs} for increasing values of $\rmse \left( \hat{y}_{adv}, y \right)$, where $\rmse \left( \hat{y}, y \right) = 2$ (solid), $3$ (dashed) and $4$ (dotted). When $\rmse \left( \hat{y}_{adv}, y \right)$ tends to infinity, the \gls{prs} converges to zero
    }
    \label{fig:prs}
  \end{minipage} \hfill
  \begin{minipage}[t]{0.65\textwidth}
   \vspace{0pt}
   \includegraphics[trim={0 0 0 2cm},clip,width=\textwidth]{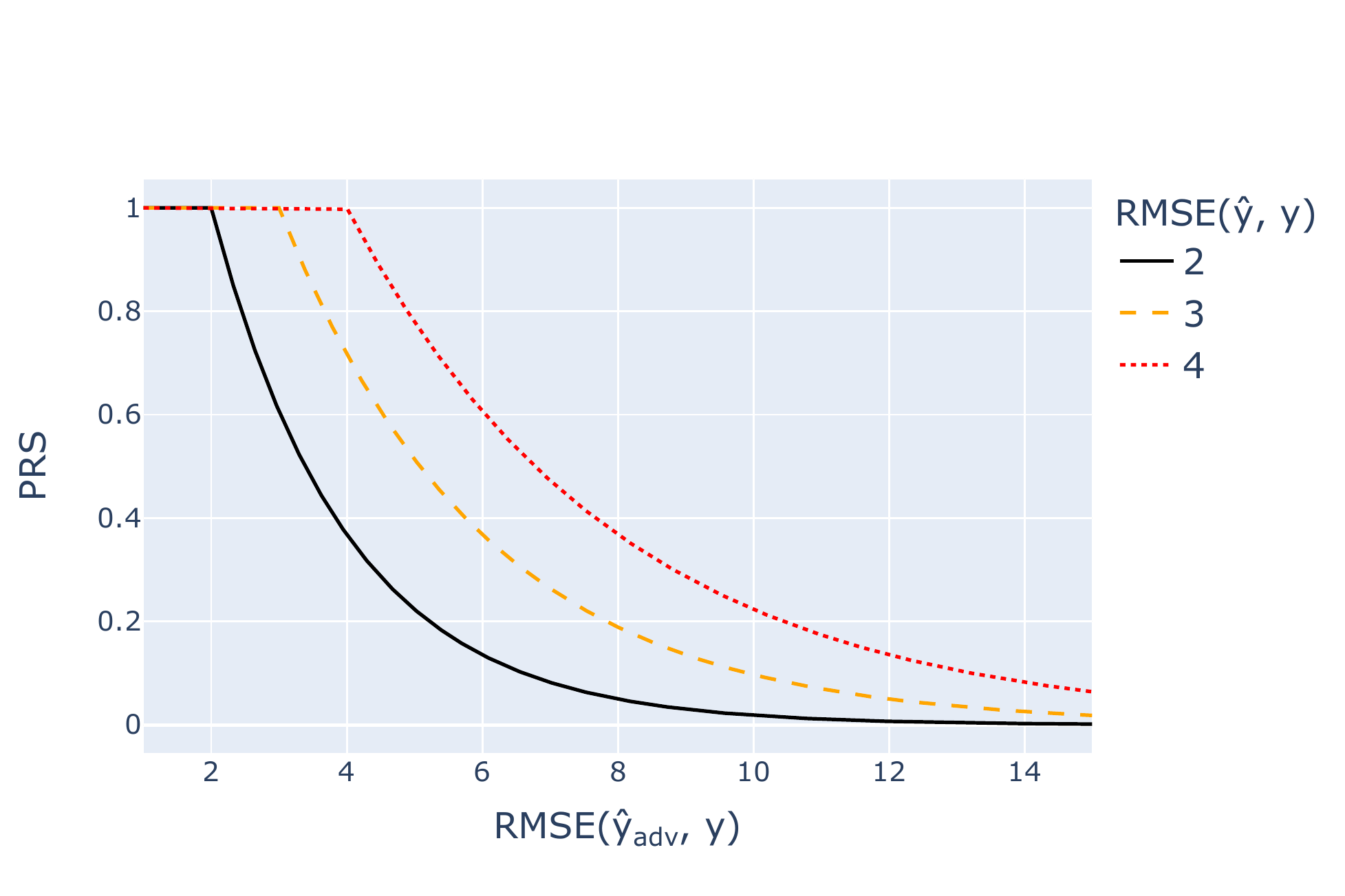}
  \end{minipage}
\end{figure}

\begin{figure}[H]
  \begin{minipage}[t]{0.32\textwidth}
  \caption{Evolution of the \gls{drs} for decreasing values of $\rmse \left( \hat{y}_{adv}, y_{adv} \right)$, where $\rmse \left( \hat{y}, y_{adv} \right) = 2$ (solid), $3$ (dashed) and $4$ (dotted). When $\rmse \left( \hat{y}_{adv}, y_{adv} \right)$ tends to zero, the \gls{drs} converges to zero as well
    }
    \label{fig:drs}
  \end{minipage} \hfill
  \begin{minipage}[t]{0.65\textwidth}
   \vspace{0pt}
   \includegraphics[trim={0 0 0 2cm},clip,width=\textwidth]{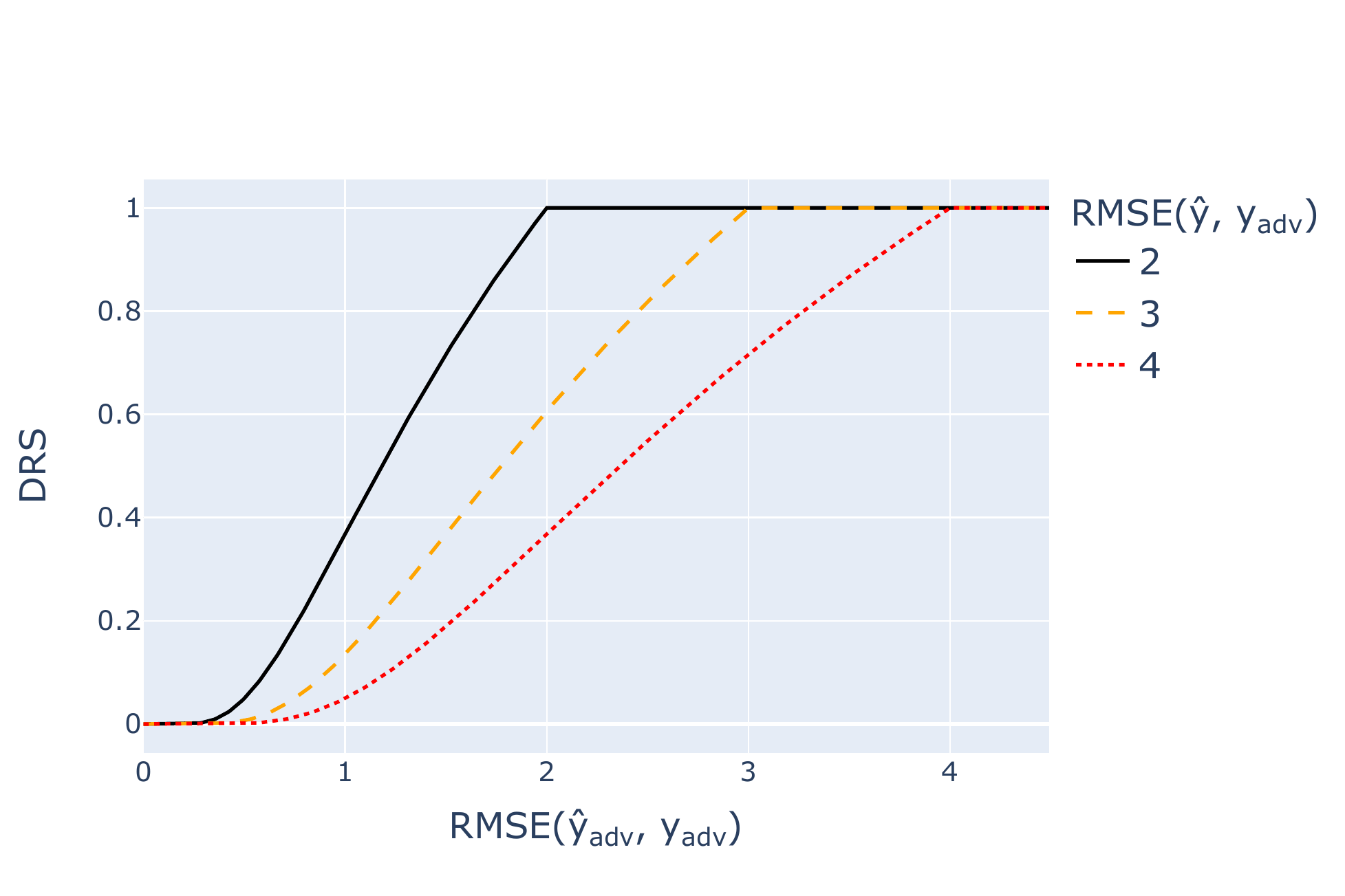}
  \end{minipage}
\end{figure}

\begin{figure}[H]
  \begin{minipage}[t]{0.4\textwidth}
  \caption{Trajectory of the $\tars _{\beta}$ with $\beta =1$ for different values of \gls{prs} and \gls{drs}. If either the \gls{prs} or the \gls{drs} take values close to zero, the value of the \gls{tars} is also close to zero. Conversely, the value of the \gls{tars} is close to one only if both the \gls{prs} and the \gls{drs} take values close to one
    }
  \label{fig:tars}
  \end{minipage} \hfill
  \begin{minipage}[t]{0.57\textwidth}
   \vspace{0pt}
   \includegraphics[trim={0 0.5cm 0 2.5cm},clip, width = \textwidth]{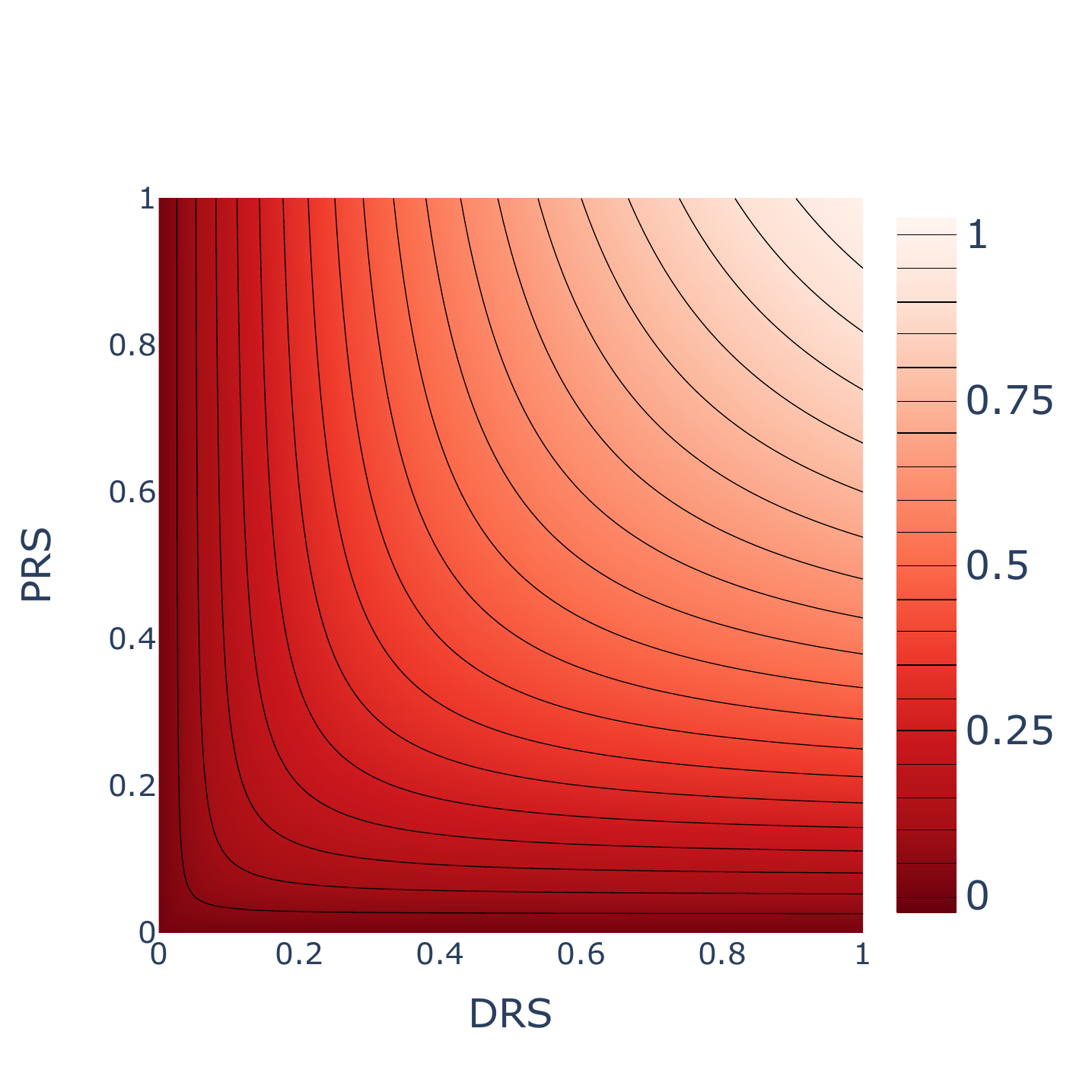}
  \end{minipage}
\end{figure}

The following figures are intended to illustrate the behavior of the three evaluation metrics \gls{tars}, \gls{drs}, and \gls{prs}. While Figure \ref{fig:prs} shows the evolution of the \gls{prs}, Figure \ref{fig:drs} demonstrates the behavior of the \gls{drs}, and Figure \ref{fig:tars} depicts the trajectory of the \gls{tars}.

\section{Data}
\label{secA1:data}
In the following, the two datasets used for the experiments in this work are described in more detail.
\subsection{Dataset on wind power generation from wind farms}
\label{subsecA1:data_wind_farm}
For the prediction of the power generation of individual wind farms, we used the publicly available\footnote{The complete data can be downloaded here: \url{https://www.dropbox.com/s/pqenrr2mcvl0hk9/GEFCom2014.zip?dl=0}} GEFCom2014 wind forecasting dataset \cite{hong2016probabilistic}. 
This dataset consists of normalized wind power measurements from 10 wind farms in Australia.
All wind power measurements were normalized to the nominal capacity of the respective wind farm and therefore took values between 0 and 1.
In addition, the dataset contains predictions of the zonal wind speed $u$ (wind parallel to latitude) and the meridional wind speed $v$ (wind parallel to longitude) at 100m above ground for the location of each wind farm. 
For simplicity, we calculated the horizontal wind speed $V_h$ at 100m above the ground from the zonal and meridional wind speeds for our experiments:
\begin{equation}
    V_h = \sqrt{u^2 + v^2}
\end{equation}
Thus, the wind power and wind speed data for the wind farms are each a univariate time series. 
The wind speed data of the wind farms were standardized separately with the z-score\footnote{The z-score is a method for normalizing a dataset by transforming its features such that they conform to a standard normal distribution with a mean of 0 and a standard deviation of 1. The z-score $z$ of an individual datapoint $x$ is calculated by subtracting the mean $\mu$ of the training dataset from the datapoint and then dividing the result by the standard deviation $\sigma$ of the training dataset, i.e. $z = \left( x - \mu \right) / \sigma$.}. 
Hence, the standardized wind speed of each wind farm had a mean of 0 and a standard deviation of 1.
The data is available for the years 2012 and 2013 with a temporal resolution of 1 hour.
For the training and hyperparameter tuning of the \gls{lstm} forecasting models, the data for each wind farm was split into a training, validation, and test dataset.
The resulting 10 training datasets each contained data from January 2012 to June 2013, with the last week of each quarter used for the corresponding validation dataset.
Thus, the training data for each wind farm spanned a total of 16.5 months, while the validation data covered 6 weeks.
The test datasets consisted of data from July 2013 to December 2013.
The individual data samples were then constructed using a one-step sliding window that moved across the hourly values.

\subsection{Dataset on wind power generation in Germany}
\label{subsecA1:data_germany}
The wind power generated throughout Germany was predicted using wind speed forecasts in the form of weather maps. 
The forecasts for horizontal wind speed at about 100m above the ground were calculated based on the zonal and meridional wind speed forecasts from the ICON-EU\footnote{\url{https://www.dwd.de/DWD/forschung/nwv/fepub/icon_database_main.pdf}} model of the \gls{dwd}. 
The wind speed forecasts\footnote{The wind speed forecasts of the \gls{dwd} in a regular latitude-longitude grid can be downloaded here: \url{https://opendata.dwd.de/weather/nwp/icon-eu/}} had an hourly temporal resolution. 
They were aggregated to a 100 × 85 grid with a spatial resolution of 10km x 10km using bilinear interpolation.
The grid covered all of Germany, and the center of the top left grid cell had the latitude 55.866 and longitude 3.071.
The historical wind power data we used as target values is real and publicly available, as are the wind speed forecasts.
Historical data on onshore wind energy generated across Germany were obtained from the website of the \gls{entso-e}\footnote{\url{https://transparency.entsoe.eu}}. 
The wind power measurements\footnote{The wind power measurements for Germany can be downloaded here: \url{https://transparency.entsoe.eu/generation/r2/actualGenerationPerProductionType/show}} were normalized by the installed wind power capacity\footnote{The installed wind power capacity for Germany can be downloaded here: \url{https://transparency.entsoe.eu/generation/r2/installedGenerationCapacityAggregation/show}} in Germany and therefore only took values between 0 and 1. 
The dataset covered the period from January 2019 to June 2021.
It was divided into 8 different subsets using blocked cross-validation.
Each subset was further subdivided into a training, validation, and test dataset.
These were chosen so that there was only a 50\% overlap between successive training datasets and no overlap between test datasets.
The time periods of the training and test datasets of the 8 cross-validation subsets are shown in Table \ref{tab:cross_validation_germany}.
\begin{table}[h]
\begin{center}
\caption{The training and test periods of the 8 cross-validation subsets of the dataset used for wind power forecasting across Germany}
\label{tab:cross_validation_germany}%
    \begin{tabular}{@{}lll@{}}
    \toprule
         \textbf{Cross-validation subset} &
         \textbf{Training} & \textbf{Test}
         \\
        \midrule
        \textbf{1} & January 2019 - June 2019 & July 2019 - September 2019 \\
        \textbf{2} & April 2019 - September 2019 & October 2019 - December 2019 \\
        \textbf{3} & July 2019 - December 2019 & January 2020 - March 2020 \\
        \textbf{4} & October 2019 - March 2020 & April 2020 - June 2020 \\
        \textbf{5} & January 2020 - June 2020 & July 2020 - September 2020 \\
        \textbf{6} & April 2020 - September 2020 & October 2020 - December 2020 \\
        \textbf{7} & July 2020 - December 2020 & January 2021 - March 2021 \\
        \textbf{8} & October 2020 - March 2021 & April 2021 - June 2021 \\
    \bottomrule
\end{tabular}
\end{center}
\end{table}
\noindent
The last four days of each month of a training dataset were used as the corresponding validation dataset. 
Thus, each training dataset spanned a total of about 5.2 months, while the validation datasets covered 24 days each.
The eight test datasets contained 3 months each.
The wind speed predictions of each subset were standardized separately using the z-score. 
Thus, the standardized wind speed predictions of each subset had a mean of 0 and a standard deviation of 1.
The individual data samples were then constructed using a one-step sliding window.

\section{Forecasting models}
\label{secA2:forecasting_models}
In the following, the two wind power forecasting models, whose adversarial robustness was investigated in this work, are described in more detail.

\subsection{LSTM forecasting model}
\label{subsecA2:wp_forecast_single}
Similar to \cite{lu2018short}, we used an encoder-decoder \gls{lstm} \cite{sutskever2014sequence} for a multistep-ahead prediction of the power generated by individual wind farms. 
This model consisted of an encoder \gls{lstm} network and a decoder \gls{lstm} network. 
First, the encoder network encoded an input sequence consisting of the wind power measurements for the past 12 hours into a latent representation. 
This latent representation was then used to initialize the hidden state and cell state of the decoder network. 
The decoder then sequentially generated a wind power forecast for the next 8 hours with a time resolution of one hour. 
Here, the decoder used the wind speed forecast of time $t$ along with the predicted wind power of the previous time $t-1$ to predict the wind power for time $t$, where $t = 1, ..., 8$. 
In the case where $t = 1$, the decoder used the real wind power measurement from the current time $t=0$ instead of a prediction. \\
The training and validation datasets of the wind farm in zone 1 of the GEFCom2014 dataset were used to tune the hyperparameters.  
The following hyperparameters of the model were optimized using the HyperBand method\footnote{HyperBand is a variation of random search that stops low-performing trials at an early stage through adaptive resource allocation and early stopping, thus speeding up the search for the optimal hyperparameters \cite{li2017hyperband}.} \cite{li2017hyperband}: number of layers, hidden size, learning rate, and length of the input sequence of wind power measurements for the encoder.
We used the asynchronous HyperBand algorithm from Ray Tune \cite{liaw2018tune} with 1000 trials and the default parameter settings. 
Only the grace period was set to 20 to avoid stopping trials too early.
After tuning the hyperparameters, the encoder network consisted of one \gls{lstm} layer with 32 neurons. 
The decoder network also consisted of one \gls{lstm} layer with 32 neurons, but followed by a dense layer with one neuron and a Leaky ReLU activation function. 
The loss function used was the \gls{mse} loss. 
As optimizer, Adam \cite{kingma2014adam} was used. 
The initial learning rate was 0.01 and was reduced by a factor of 0.1 each time the validation loss did not improve over 10 epochs, using a learning rate scheduler. For this purpose, PyTorch's \cite{paszke2019pytorch} ReduceLROnPlateau learning rate scheduler was used with the default parameter settings. 
The maximum number of epochs was constrained to 100.
Preliminary experiments have shown that this number is sufficient for convergence of the model's training.
In addition, early stopping was used to stop the training as soon as the validation loss did not improve within 15 epochs.
Here, the EarlyStopping callback from PyTorch Lightning \cite{falcon2019pytorch} was used with the default parameter settings. 
Only the patience parameter was chosen as 15 epochs, since this improved the model's performance in preliminary experiments.

\subsection{CNN forecasting model}
\label{subsecA2:wp_forecast_germany}
A new approach for forecasting the generated wind power in large-scale regions was proposed by \cite{bosma2022estimating}. 
In this approach, the problem of wind power forecasting is divided into two distinct subproblems, each of which is solved separately. 
The first step consists of generating very accurate weather forecasts using a suitable weather prediction model. 
The second step then consists of generating the wind power forecast using the weather forecasts. 
For this purpose, a separate power estimation model is applied to estimate the wind power for a future point in time using the predicted weather maps for that point in time and previous points in time. \\
We used this approach in order to make an 8h forecast with one-hour resolution for the wind energy generated throughout Germany. 
To make a forecast for time $t$, the model received a stack of 5 weather maps as input. 
These consisted of the forecasts for the horizontal wind speed at 100m above ground level for the 5 hours leading up to the estimation time, i.e., points in time $t-4, ..., t$.
Here, the wind speed prediction of each point in time represented a separate channel. Thus, the dimension of the input data for a prediction for time $t$ was 5 × 100 × 85 (channels × pixel height × pixel width). 
For estimating the wind power based on the 5 weather maps, we used a ResNet-34 \cite{he2016deep}, followed by a dense layer with one neuron and a Leaky ReLU activation function in the output layer. 
This model was then sequentially applied to the input data and estimated the generated wind power step-by-step for points in time $t=1, ..., 8$. 
For training the model, \gls{mse} loss was used. 
As optimizer we used Adam. 
The maximum number of epochs was limited to 100. Preliminary experiments have shown that this number is sufficient for convergence of the model's training.
The initial learning rate was 0.001, which is the default value of PyTorch's \cite{paszke2019pytorch} Adam optimizer.
It was reduced by a factor of 0.1 each time the validation loss did not improve within 10 epochs. 
For the \gls{cnn} model, we used early stopping and the ReduceLROnPlateau learning rate scheduler in the same way as for the \gls{lstm} model, see Section \ref{subsecA2:wp_forecast_single} for a detailed description.




\end{document}